# An Unsupervised Deep Explainable AI Framework for Localization of Concurrent Replay Attacks in Nuclear Reactor Signals

Konstantinos Vasili*, Zachery T. Dahm, Stylianos Chatzidakis

**Affiliation:** School of Nuclear Engineering, Purdue University, West Lafayette, IN 47907

***Corresponding author:** vasilik@purdue.edu

**Abstract**

Next-generation advanced nuclear reactors are expected to be smaller in size and power output, relying extensively on fully digital instrumentation and control systems. These reactors will generate a large flow of information, conveying simultaneously various non-linear operational states. Ensuring data integrity against deception attacks is becoming more and more important for networked communication, cyber-physical control systems, and a requirement for safe operation. Current efforts to address replay attacks, a special kind of deception cyber event, almost universally focus on watermarking or anomaly detection approaches without further identifying the root cause of the anomaly. In addition, they rely mostly on synthetic data with uncorrelated Gaussian process, measurement noise and full state feedback, or are limited to univariate signals, signal stationarity, linear quadratic regulators, or other linear time-invariant state-space models, which may fail to capture unmodeled system dynamics. In the realm of regulated nuclear cyber-physical systems, additional work is needed on characterization and explainability of replay attacks using real data. Here, we propose an unsupervised explainable AI (XAI) framework based on a combination of an autoencoder and customized windowSHAP algorithm to fully characterize real-time replay attacks, i.e., detection, source identification, timing and type, of increasing complexity during a dynamic time-evolving reactor process. The proposed XAI framework was benchmarked on real-world datasets from Purdue's nuclear reactor, with up to six signals concurrently being replayed. In all cases, the XAI framework was able to detect and identify the number of signals being replayed and the duration of the falsification with 95% or better accuracy.

## Introduction

Currently, there is a growing trend in the nuclear industry toward advanced reactor designs, such as microreactors and small modular reactors. To make these designs cost competitive and operationally flexible, digital instrumentation and control, supported by wired or wireless communication networks, will need to be implemented. These systems will offer a vast amount of information, and promise centralized or semi-autonomous control, real time monitoring, prognostics, diagnostics, and unattended operation [1], [2], [3], [4]. Despite the numerous advantages of digital instrumentation and control, new challenges emerge, such as the near impossibility of maintaining constant human surveillance over potential anomalies within the digital system [5]. Moreover, the interconnectivity through networked flow of information makes them vulnerable to cyberattacks. Unlike conventional information technology (IT) systems, where the main focus is the protection of network data, cyber events on critical, networked control infrastructure could affect not only the data but also physical and control processes through a variety of means, potentially causing economic loss, (e.g., due to unnecessary shutdown of the reactor) or equipment damage [6] [7]. The nuclear industry has been a target for cyber threats for many years [8][9]. It is worth noting the infamous Stuxnet malware detected in 2010, which affected the digitized industrial control system of a nuclear facility [10] [11] along with other non-nuclear facilities around the world. Other examples include the Korea Hydro and Nuclear Power incident in 2014 [12], the Davis-Besse NPP event in 2011 [13], and the Browns Ferry NPP Unit 3 incident in 2006 [14]. These cyber incidents suggest that the introduction of new vulnerabilities to nuclear systems, previously absent due to analog-base operation, has become a growing concern [15]. Consequently, there is increasing interest in developing detection methods for cyberattacks targeting nuclear and industrial control systems.

In the broader literature, newly developed methods have explored resilient and attack-tolerant control strategies to improve the robustness of networked and hybrid systems under cyber-physical uncertainties or other disturbances. For instance, these studies [16], [17], [18] are proposing consensus and coordination algorithms to ensure stable operation in distributed multi-agent systems under asynchronous communication or competition–cooperation interactions. Similarly, composite anti-disturbance control methods and event-triggered feedback schemes have been proposed to maintain system stability in the presence of actuator attacks and network perturbations [19]. More recently, researchers explored the integration of artificial intelligence (AI) into nonlinear control design for hybrid energy and vehicular systems, where AI-based controllers are used to optimize power conversion and energy management performance [20], [21]. These studies offer valuable insights into enhancing the resilience and performance of complex control systems, however their focus is primarily on maintaining or improving system behavior through model-based control mechanisms. A complementary equally important approach for intelligent systems resilience is to detect and characterize the effects of cyber-induced deviations that are already present in the system.

Towards this direction, Teixeira et al. [6] present an overview of cyber events that could compromise networked control systems and define a three-dimensional cyber-physical attack space comprising knowledge, disclosure, and disruption resources. The main properties of an attack include violation of data availability, integrity, and confidentiality. The characteristic example of a data availability violation is a Denial of Service (DoS) attack [22] [23] where an adversary prevents data from reaching its destination. Examples of data integrity violations include deception attacks such as False Data Injection (FDI) [24] [25] and replay attacks [26] [27]. FDI attacks have received attention from researchers and a comprehensive review against modern power systems can be found in [28]. Unlike an FDI attack, which introduces conceptually erroneous or inconsistent data into the system, a replay attack involves intercepting a valid sequence of past measurements and resending them at a later time to deceive the operators about the current system state [6] [21]. This makes replay attacks more difficult to detect and has consequently resulted in fewer studies in the area. During Stuxnet, though not primarily a replay attack, recorded normal data were replayed to the monitoring system to create the illusion that the system was operating normally, while an even more sophisticated attack was ongoing.

The techniques proposed in the literature for identifying replay attack events can be broadly categorized into model-based and data-driven. Model-based methods rely on system knowledge to construct a representation of the control system [26] [27] [29] [30] [31], with one of the most common techniques being watermarking. This detection approach involves adding small random perturbations (watermarks), unknown to the attacker, to the control input in order to authenticate the system. In one of the earliest works in the field, Mo and Sinopoli, (2009) [26] investigated a discrete, linear time invariant, Gaussian control system for anomaly detection. They proved that the residues from the controller and the predicted measurements from a Kalman filter follow a Gaussian distribution. Subsequently, a $\chi^2$ failure detector was used for intrusion detection by adding random noise to the signal and comparing the difference to a threshold in a predefined window. Yao and Smidts, (2020) [27] extended this approach by adding a second $\chi^2$ detector to include classification of the anomaly into FDI or other types. Similar watermarking techniques are proposed in more recent works [29] [30] [31]. The authors in [31] perform replay attack detection with a deterministic event-based physical watermark technique. While successful, the above studies relied on linearized mathematical models, uncorrelated Gaussian noise, and required specific parameterizations to represent the non-linear dynamics. These factors pose practical limitations, as such information is not always available. Additionally, a tradeoff exists between detection and system performance, which can become a critical concern when the system is compromised. Moreover, most studies assume a priori knowledge of the compromised signal, signal stationarity and rely on synthetic data for proof of concept, which may limit their applicability under real world operational conditions.

Alternatively, data driven techniques have also been proposed in various domains [32] [33] [34] [35] [36]. Replay attacks, as a specific subclass of cyber threats, fall into the general field of time series anomaly detection, and tools from this field are typically employed to address the challenge. Of relevance to our work, Gargoum et al. (2025) [34] propose a data driven, two stage framework for anomaly detection and verification. The approach combines matrix profile-based change-point detection with a convLSTM architecture to distinguish normal from replayed sensor measurements and introduces spatio-temporal features to subsequences of the time series flagged as replay attacks to verify the detection. Zhao et al. (2024) [35] proposed a data driven dynamical delay estimation method that detects replayed measurements by identifying temporal inconsistencies between actual and injected signals. Despite the novelty, these methods appear to be applied to a single signal and do not address scenarios involving multi-sensory or multivariate environments, where replay attacks may target arbitrary combinations of signals. Furthermore, even if separate frameworks are deployed to monitor each signal, these approaches would fail in scenarios where anomalies are contextual, flagged by the absence of expected inter-signal correlations rather than deviations in individual signal patterns. In multivariate settings, AA Elsaeidy et al. (2020) developed a Deep Learning (DL) pipeline for detecting replay attacks in smart cities [32]. A multi-sensor infrastructure was designed to collect multivariate data, and several DL architectures, including Convolutional Neural Networks (CNNs), were trained to solve a binary classification task, distinguishing normal from replayed time series. Wang et al. [36] explored a Long Short-Term Memory (LSTM) framework for detecting FDI and replay attacks in multivariate times series (MTS) data for an industrial control process. The framework successfully identified replayed sequences by comparing predictions with a predefined error threshold. However, the study considered a scenario in which all signals were falsified, creating clearly distinguishable patterns, while unable to identify specific affected signals or determine the timing and duration of the attack. Other approaches propose clustering based statistical methods [37] [38] and classical ML techniques such as SVM [39]. Although suitable for their specific applications, these methods do not consider the time component and therefore are not suitable for real time monitoring.

Despite the success of the above frameworks, developing methodologies for fault and cyber event characterization in nuclear reactors remains a major engineering challenge, as such systems must be able to detect, localize, and classify diverse cyber-physical scenarios without compromising reactor safety [40]. Furthermore, the vast amount of information provided by such systems increases the attack space making it impractical to monitor individual signals with dedicated models by applying existing techniques. To address some of those limitations, Dahm et al., (2025) [33] added interpretability to an LSTM predictive framework for detecting replay attacks in MTS data from the PUR-1 nuclear reactor. The framework, successfully demonstrated using real data from PUR-1, identified nonstationary, concurrent replay attack scenarios of increasing complexity by extracting Shapley values to interpret the difference between a critical signal prediction and the actual signal value. However, the framework relies on the prediction of a single critical signal, not known a priori to the adversary, which limits its applicability when the critical signal is itself the target of a possible attack.

With advancements in computational units observed in recent years, significant attention has been dedicated to anomaly detection using AI/ML [41]. Among various approaches, unsupervised methods using Autoencoders (AE) have shown strong performance, especially in systems where abnormal events are rare. A promising research direction explores the combination of deep architectures with Explainable AI (XAI) algorithms to provide interpretable insights into deviations from normal patterns [42] [43] [44]. However, despite its importance, the temporal dimension in XAI-based analysis has received limited attention. In this work, we aim to fill this gap by combining and expanding two individually proven techniques into a unified, robust, and efficient framework for cyber anomaly detection. Motivated by the increasing number of cyber threats observed in the nuclear industry, the ongoing research on networked reactor designs, and recent advancements in AI/ML/XAI-based anomaly detection methods, this paper addresses the above limitations by proposing a data driven, unsupervised, deep explainable AI (XAI) framework with the following key contributions, where comparisons with existing literature is provided in Table 1.

1. A real-time replay attack detection approach is proposed for monitoring nonstationary, multivariate nuclear reactor signals with nonlinear and noisy characteristics during a dynamic reactor operation.
2. An optimal reconstructive modeling of a research reactor is employed in combination with a modified XAI algorithm to enable characterization, time of occurrence and duration of signal-level replay attack scenarios of increasing complexity. Six concurrent replay attack scenarios are evaluated, in which the framework successfully detects the number of replayed signals, the start time and duration of each attack.
3. The proposed framework extends beyond binary classification by quantifying feature importance across temporal and feature dimensions. It identifies a general condition that triggers deviation from expected reactor behavior and further evaluates each signal's contribution by computing Shapley values over time and feature space. Accurate background data are used, which are dynamically updated to account for the dynamics of the reactor. This approach makes the localization of falsified signals possible, even if multiple signals are replayed.
4. An analysis of the Shapley values distribution during the attack is performed, demonstrating that the framework can effectively characterize the specific attack through the evaluation of multi-signal deviations from expected behavior.
5. The proposed approach was benchmarked against real-world data from Purdue's nuclear reactor (PUR-1) and can be extended to any control system with predictable operational modes making the structure of the proposed framework generalizable to other domains.

Table 1: Comparison of proposed framework with existing literature

| Framework | Real-time detection | Time of occurrence | duration | Multianomaly detection - localization | Non-stationary signals | Generalizable |
|---|---|---|---|---|---|---|
| [32] | Yes | No | No | No | Not specified | No |
| [33] | Yes | Yes | Yes | Yes | Yes | No |
| [34] | Yes | Yes | Yes | No | Yes | No |
| [35] | Yes | Yes | Yes | No | Not specified | No |
| [36] | Yes | Yes | Yes | No | Yes | No |
| This paper | Yes | Yes | Yes | Yes | Yes | Yes |

The remainder of the paper is organized as follows. Section II presents the use case, the data extraction and the simulation of the attack scenarios. Section III introduces the methodology. Section IV describes the implementation of each module in the methodology. Section V details the results of the proposed framework, and lastly Section VI summarizes the conclusions of the research and highlights future work to further develop the proposed cyber-attack detection system.

# Use case

### Scenario

The use case involves injecting multiple replay attacks into different sensors using data recorded during a past steady-state condition. This experimentally simulates an adversary attempting to deceive the operator into believing that the system remains at high power during a forced shutdown process. The event starts when the SCRAM button is activated. This cuts the magnet current to the control rods, which are then rapidly inserted into the reactor core. The control rods are made of neutron absorbing material and immediately terminate the reactor's heat production process. From that point, the reactor power is due to residual heat generated from precursor concentrations. In a normal process, the sensor readings exhibit rapid changes, as shown in Figure 1 (left). The conceptual cyber scenario is illustrated in Figure 1 (right), where the green line indicates the SCRAM activation, the red line represents the expected system response, and the blue line shows the observed behavior under the cyber event.

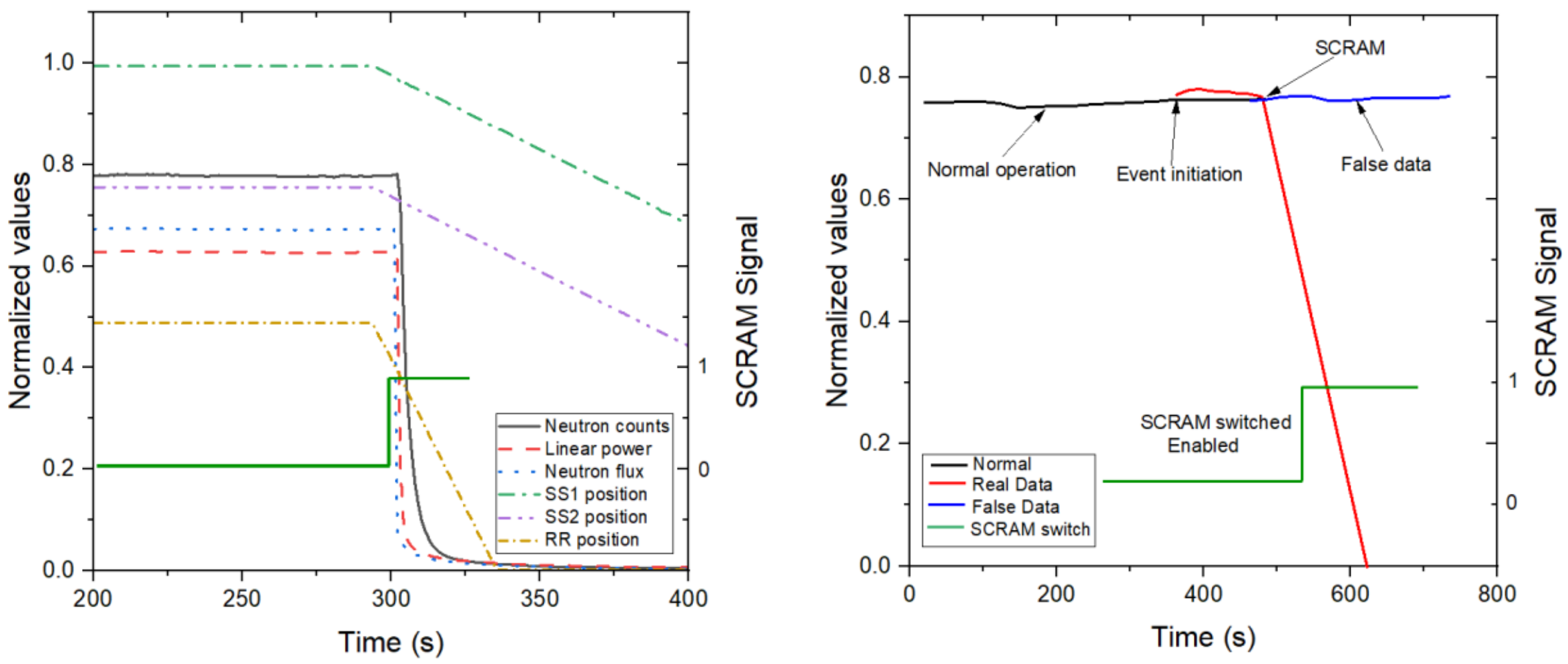

Figure 1: Variation in neutron counts following a SCRAM process (left). Cyber event initiation represented by the blue line where the shutdown process is not followed by decreasing values (right).

To explore the effectiveness of the proposed framework under this scenario, replay attack events of increasing complexity were experimentally simulated. A set of *n* sensor measurements from a control system can be represented as multivariate time series data in the following form:

$$X(t) = [x_1(t), x_2(t), \dots, x_n(t)]$$

Where each element $x_i(t)$ represents the reading from the *i-th* sensor at time t. Under the assumption that the attacker has access to the system, the replay attack is carried out in two stages, as described in relevant literature [34], [45]. In the first stage, the system is intercepted, and an unknown number of signals are recorded over an unknown time interval *I*. The recorded set can be represented as:

$$\tilde{X}(t) = [\tilde{x}_1(t), \tilde{x}_2(t), \dots, \tilde{x}_k(t)], \quad t \in I$$

The time interval $I$ is defined as: $I = [t_{start,}, \ t_{start} + T]$,

Where $t_{start}$ denotes the start time of the recording, T is the duration, and $k$ is the number of signals recorded.

In the second stage, the attacker injects the recorded set $\tilde{X}$ at a later time interval defined as: $I_{attack} = [t_{attack}, \ t_{attack} + NT]$, where $t_{attack}$ denotes the attack start time and N represents the number of times the recorded set is repeated. If $N = 1$, the attack can be regarded as a delay attack. In our use case, $k$ ranges from one and increases to six signals concurrently replayed: neutron counts (ch.1), reactor power (ch. 3), neutron flux (ch. 4), the regulating rod (rr-position), the SS1 rod, and the SS2 rod. It is assumed that these signals will be primarily a target for a cyber-attack, as they are typically monitored by operators on the control console. The recorded time interval $I$ is selected during a steady state condition prior to the SCRAM event. The duration T is set to 20 seconds of steady state operation, $t_{attack}$ is set immediately after the SCRAM button is pressed, approximately around 300 s, and N is set to 5, resulting in 100 seconds of replayed signals covering the entire SCRAM event (Figure 2). The selection of the above parameters (timing, duration and number of signals) can produce varying effects, resulting in different attack scenarios which may mislead the operator about the true reactor condition. Depending on the operator's response, the potential consequences may include significant economic losses or even threats to human safety, therefore the significance of correctly and promptly identifying the attack is evident. After the attack, all signals return to normal values, corresponding to low sensor readings caused by the large negative reactivity inserted by the control rods.

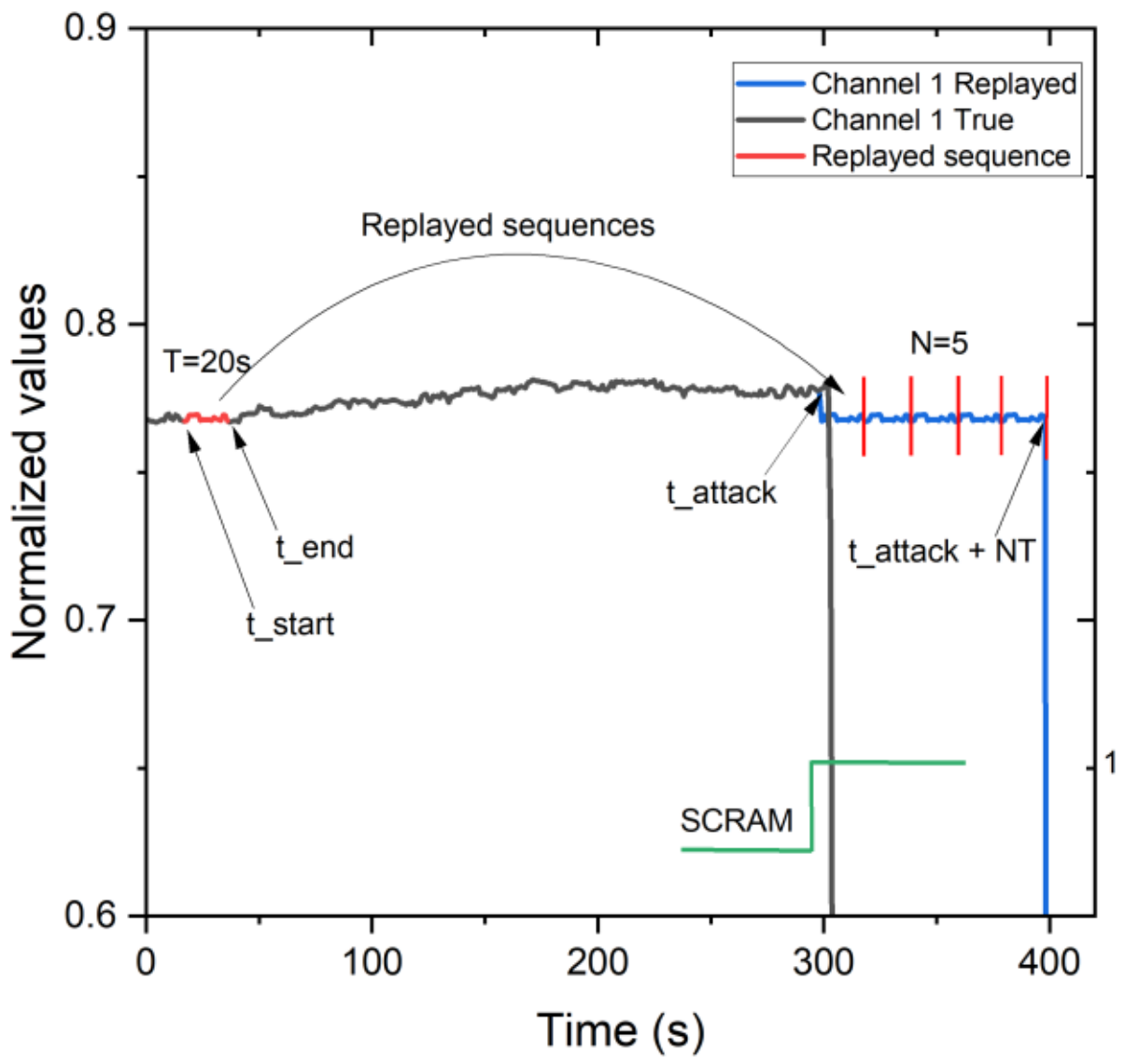

Figure 2: Example of a replay attack scenario for Channel 1. Applying the same process to additional signals simulates concurrent replay attacks of increasing complexity.

After a SCRAM process is initiated, all control rods are inserted into the reactor core over a few seconds. Figure 1 shows that the rr rod appears to insert after approximately 40 seconds, while the SS1 and SS2 rods take roughly 300 seconds to reach full insertion. This delay is primarily attributed to the tracking of the control rod drives rather than the rods themselves which have disengaged from the drives, producing a distinct pattern that allows the operators to identify a SCRAM process on the console.

**PUR-1**

Real reactor datasets are not readily available in the literature, and most studies in the nuclear domain rely on simulated or synthetic data that lack physical or artificial inconsistencies, such as noise, null values, and outliers, resulting in nearly ideal datasets for data-driven testing [46] [47] [48] [49] [50]. Here, we leverage Purdue's University research reactor PUR-1 (Figure 3)  [51] [52]. PUR-1 is a pool-type reactor licensed to operate up to 10 kWth continuous power and is the only reactor in the United States with a fully digital instrumentation and control system, which allows for real-time data access and remote monitoring. A digital twin [53] enables real-time data collection of more than 2000 parameters per sampling frequency, through a monitoring workstation (Figure 3), supported by a communication network that transmits the data.

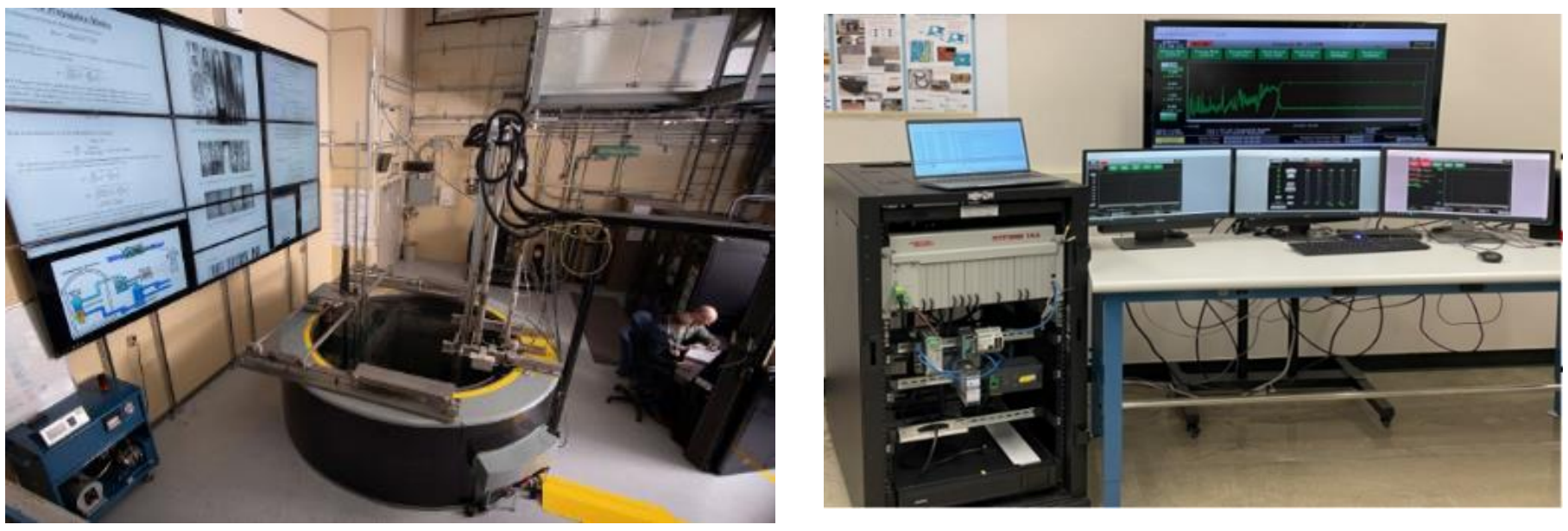

Figure 3: PUR-1 facility (left). Digital twin away from reactor building with no direct line of sight (right).

PUR-1 is only one component of a larger cyber-physical system composed of five layers, which include additional elements such as equipment, controllers, network connections and communication protocols (Figure 4) [52]. Layers 0 through 3 are located inside the PUR-1 facility, while Layer 4 is located in a separate building with no direct line of sight to the reactor. The reactor's operational dynamics can be viewed as a multidimensional temporal sequence of sensor readings generated from Layers 0 to 3. Data collection, analysis, and ML training and testing are performed at layer 4.

A fundamental requirement for any cyber event, including FDI and replay attacks, is that the intruder has successfully gained network access. For our use case, we assume that the intruder has *a priori* access at Layer 1 (but not at Layer 0) and can consequently record selected signals through an entry point and replay them at a later time. We further assume that the communication network is reliable and does not introduce disruptions such as packet losses or delays that could affect the transmitted data.

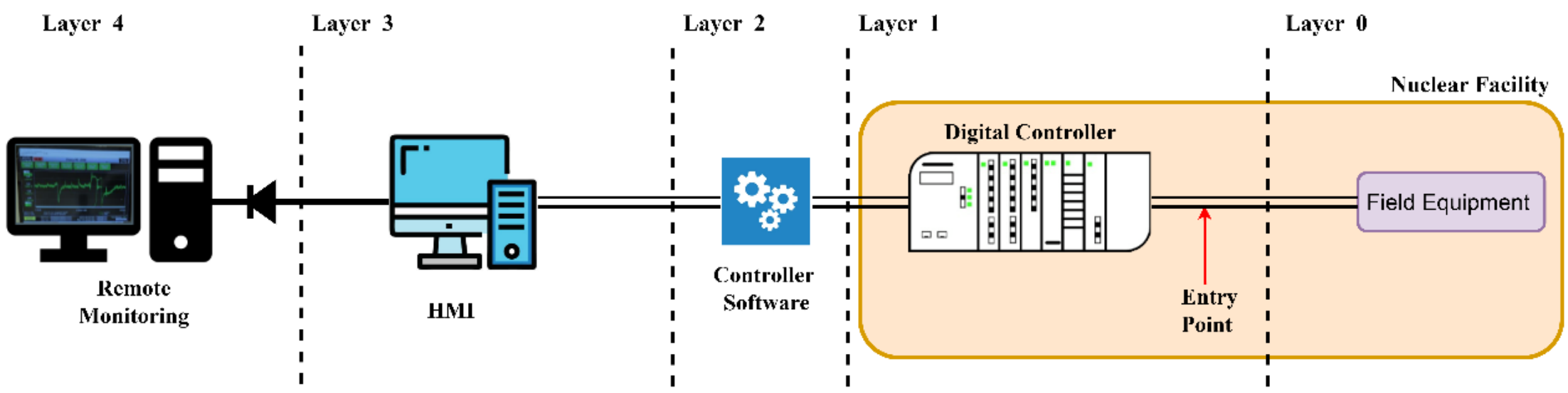

Figure 4: Cyber physical system architecture for PUR-1 cybersecurity testing.

Figure 5 presents the architecture developed to inject replay attacks and collect the resulting falsified data [52]. The architecture consists of two sections: (i) the Remote Location (left), where an attacker's PC connects to the monitoring workstation via a network switch to enable the injection of false data; and (ii) the Control Room (right), which hosts the real-time reactor data network, including controllers, sensors, human-machine interfaces (HMIs), operational processes, and incoming signals used for decision-making [33], [52].

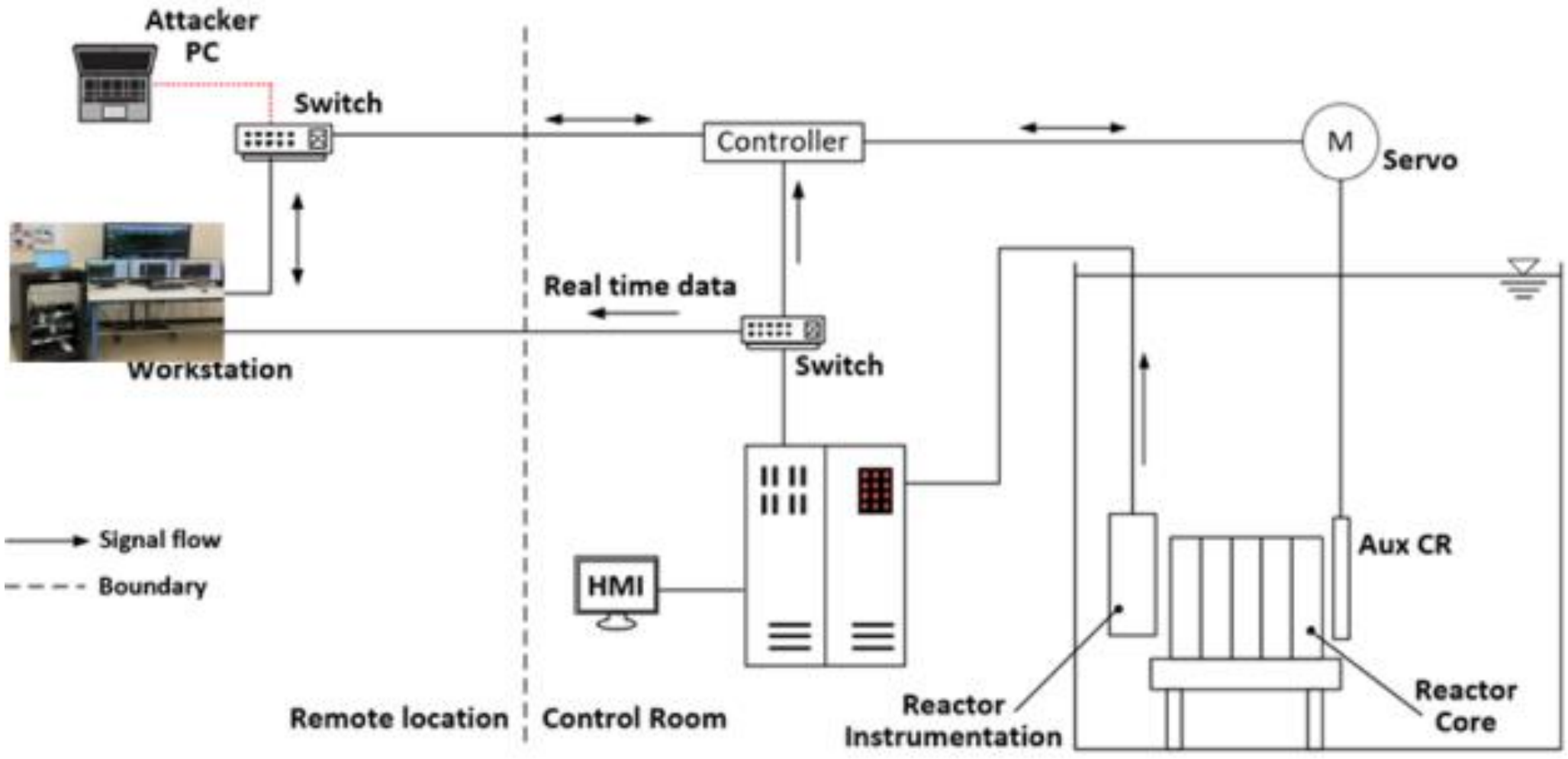

Figure 5: Architecture for injecting false data.

**Data**

The data is extracted from the digital twin workstation (Figure 3) in tabular form, capturing either the full set or a selected subset of available signals. Each row corresponds to a one-second sampling frequency, and the columns represent individual signals, collectively forming an MTS dataset that reflects the reactor's operational state at each time step. The high dimensionality of the dataset, which exceeds 2,000 signals, leads to increased computational costs and potential performance concerns. Consequently, reducing the number of signals used for model training is necessary. We identified and selected nine key signals that most accurately represent the reactor's functionality. The selected signals are listed in Table 2.

The operational data collected is subject to various artifacts, including fluctuations, outliers, null values, and noise. These inconsistencies arise from several sources, such as limitations or imperfections in the data-collection instruments and the inherent complexity of the monitoring system. Additionally, variability in human operation and decision-making contributes to data heterogeneity, as different operators may respond differently under similar conditions, a behavior reflected in the recorded data. The frequency and nature of these artifacts can also vary with the reactor's operational mode, whether during shutdown or any other active operation.

Table 2: Signals used for training, testing and falsification

| Feature | Units | Description |
|---|---|---|
| Neutron counts | 1/s | Neutron counts per second as measured by channel 1 |
| Linear power | % | Reactor power as measured by channel 3 |
| Neutron flux | % | Neutron flux as measured by channel 4 |
| ss1-position | cm | Distance an electromagnet has removed SS1 from the core |
| ss2-position | cm | Distance an electromagnet has removed SS2 from the core |
| rr-position | cm | Distance an electromagnet has removed RR from the core |
| rr-active-state | - | Categorical with three discrete values (insert-withdraw-steady) |
| ss1-active-state | - | Categorical with three discrete values (insert-withdraw-steady) |
| ss2-acrtive-state | - | Categorical with three discrete values (insert-withdraw-steady) |

A total of 47 full power cycle datasets were collected and used for the initial training of the algorithms. Each dataset contains approximately 5700 seconds (1.5 hours) of reactor operation, covering startup through shutdown and including transient and steady state conditions across various power levels (Figure 6). Additionally, 24 reactor SCRAMs, each approximately 800 seconds in duration, were collected to capture SCRAM events occurring at different power levels (Figure 7).

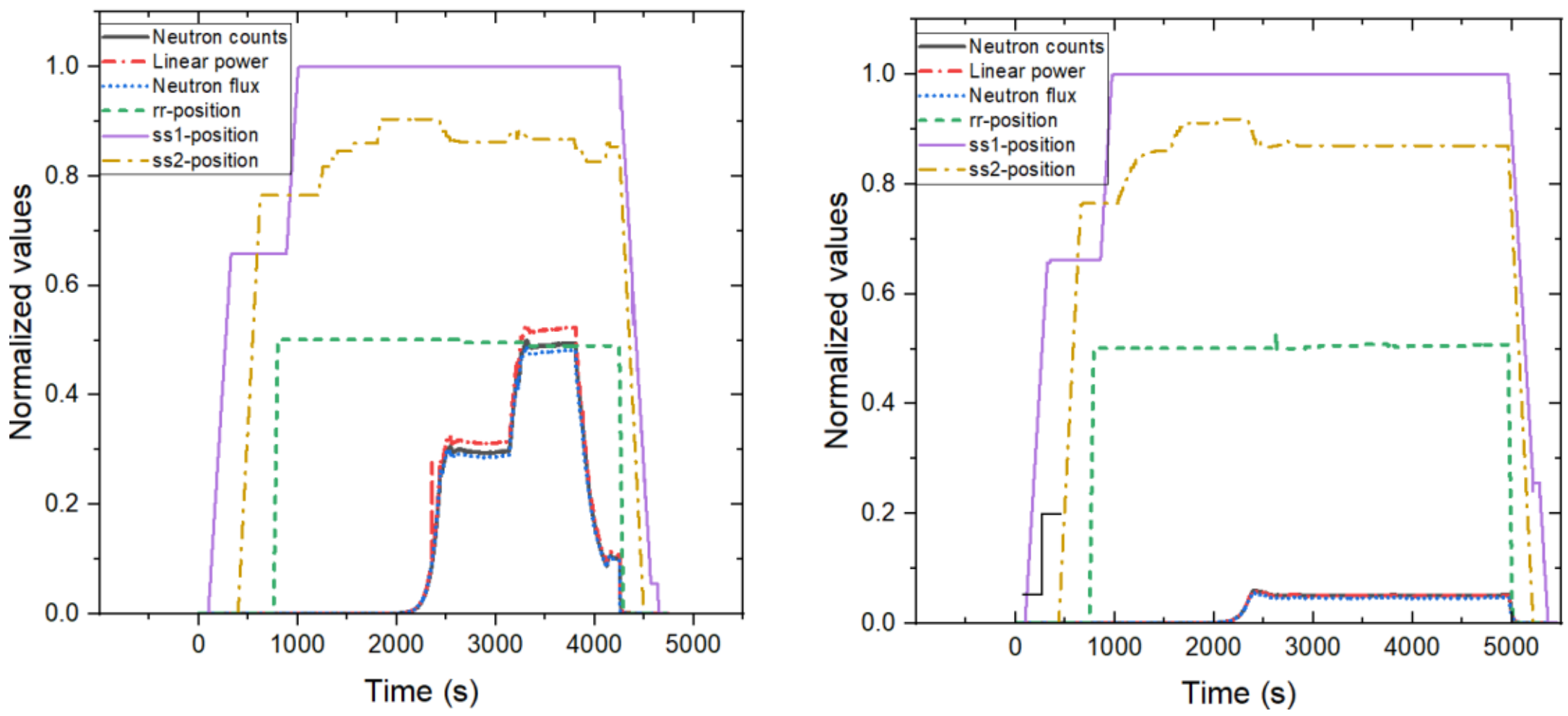

Figure 6: Examples of signal variations across different full power cycles of PUR-1 operation.

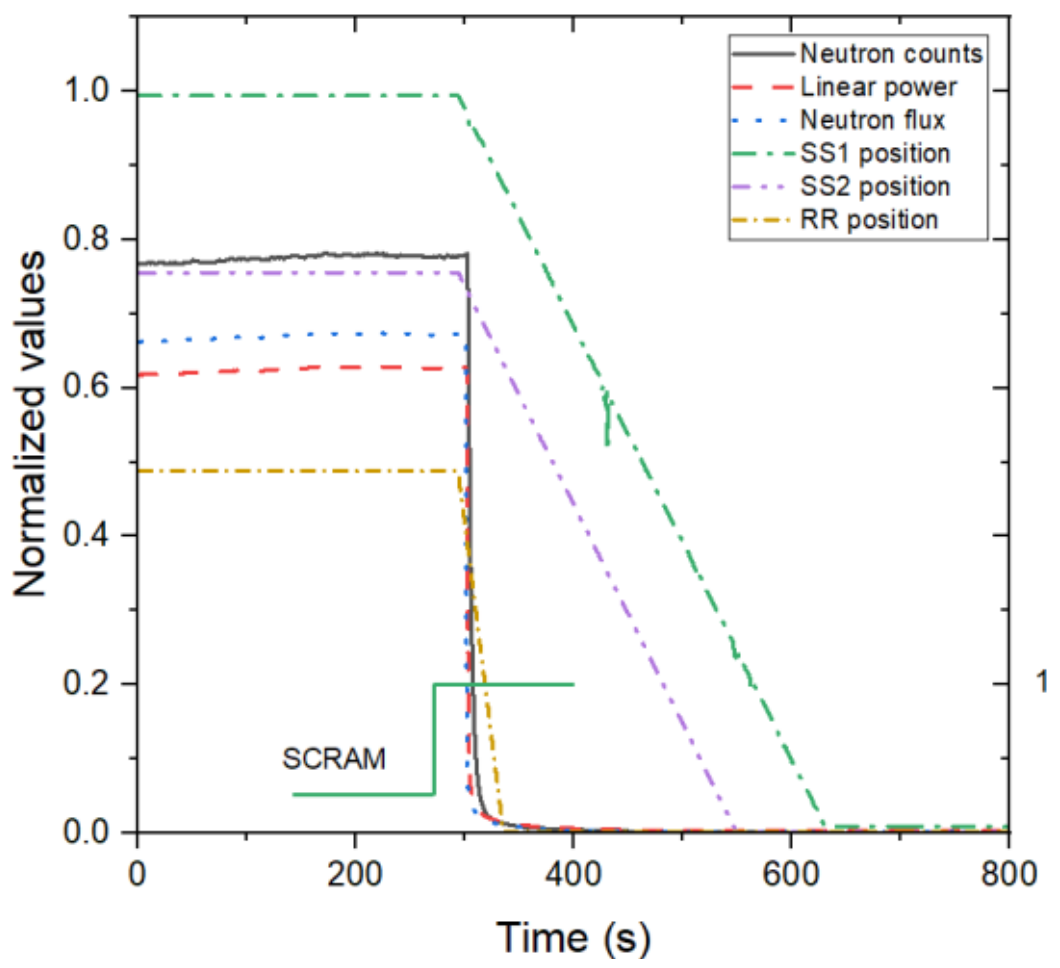

Figure 7: Expected signal behavior during a SCRAM event.

Following the process described in the scenario section, six replayed datasets were generated, each involving an increasing number of falsified signals:

1. Replay#1: Replay attack of one signal (ch.1)
2. Replay#2: Replay attack of two signals (ch.1, ch.3)
3. Replay#3: Replay attack of three signals (ch.1, ch.3, ch.4)
4. Replay#4: Replay attack of four signals (ch.1, ch.3, ch.4, RR)
5. Replay#5: Replay attack of five signals (ch.1, ch.3, ch.4, RR, SS1)
6. Replay#6: Replay attack of six signals (ch.1, ch.3, ch.4, RR, SS1, SS2)

Out of the nine signals listed in Table 2, the three control rod active states remained unchanged across all cases to allow the algorithm to detect deviations from the normal case. Figure 8 presents the six simulated datasets. All signal values were normalized to account for signal variability and ensure consistent visualization. Each dataset is divided into two panels: The left side displays the full 800 seconds of the SCRAM event under the replay attack. The right side presents the same condition restricted to the time frame in which the SCRAM is happening (300-400 seconds).

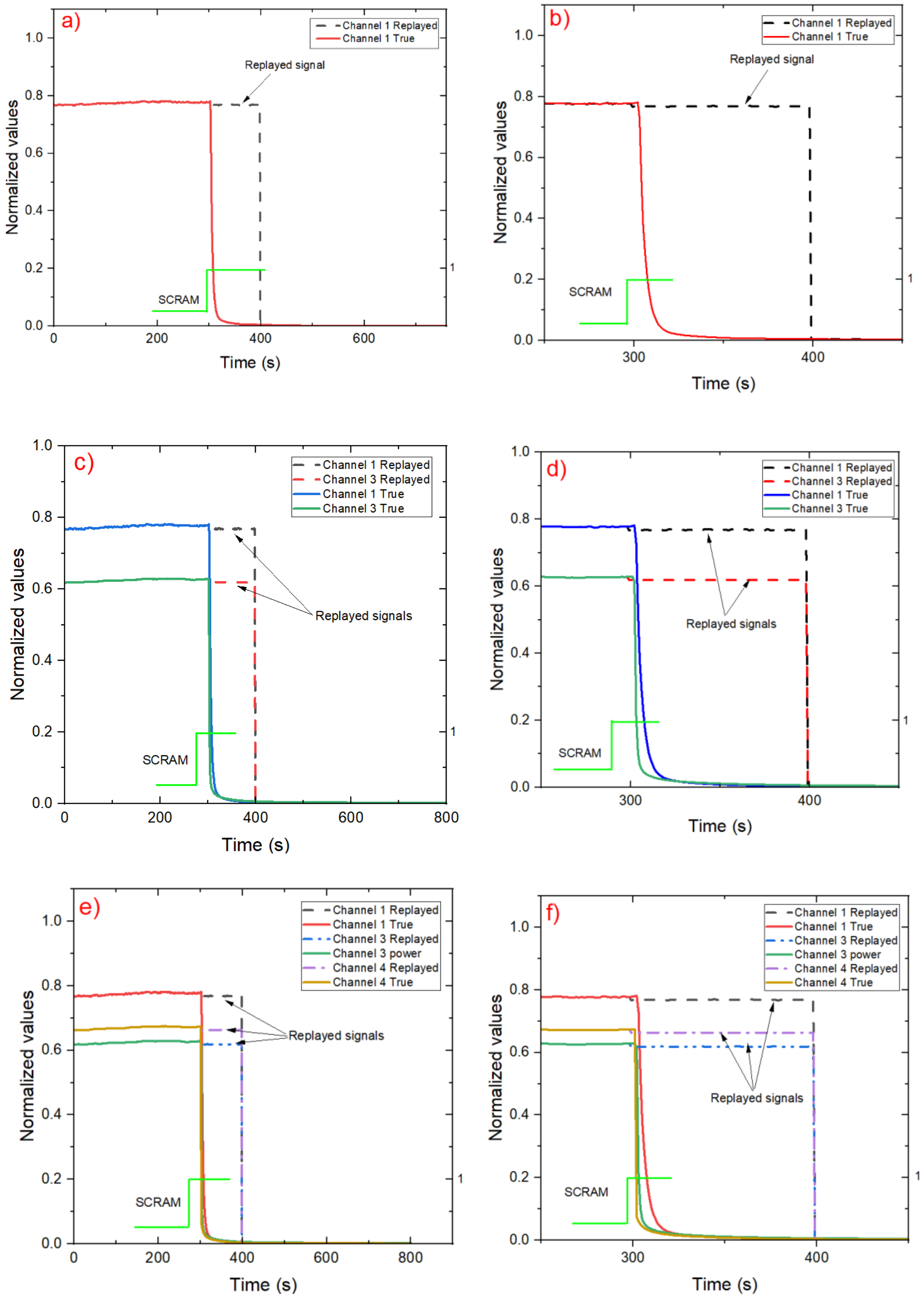
a)
Channel 1 Replayed
Channel 1 True
Replayed signal
SCRAM
Normalized values
Time (s)
b)
Channel 1 Replayed
Channel 1 True
Replayed signal
SCRAM
Normalized values
Time (s)
c)
Channel 1 Replayed
Channel 3 Replayed
Channel 1 True
Channel 3 True
Replayed signals
SCRAM
Normalized values
Time (s)
d)
Channel 1 Replayed
Channel 3 Replayed
Channel 1 True
Channel 3 True
Replayed signals
SCRAM
Normalized values
Time (s)
e)
Channel 1 Replayed
Channel 1 True
Channel 3 Replayed
Channel 3 power
Channel 4 Replayed
Channel 4 True
Replayed signals
SCRAM
Normalized values
Time (s)
f)
Channel 1 Replayed
Channel 1 True
Channel 3 Replayed
Channel 3 power
Channel 4 Replayed
Channel 4 True
Replayed signals
SCRAM
Normalized values
Time (s)

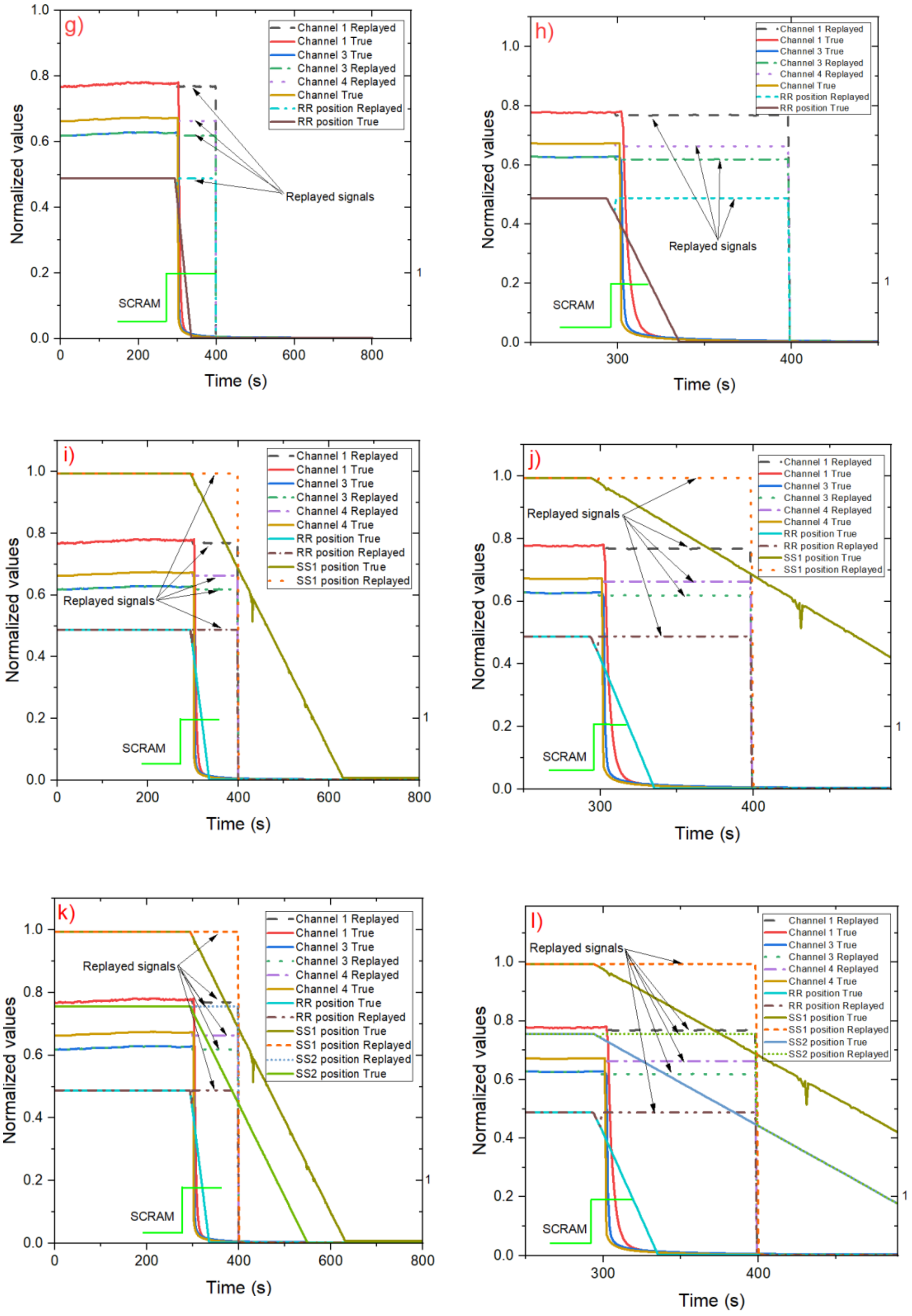

Figure 8: Replay attack one signal (a-b); Replay attack two signals (c-d); Replay attack three signals (e-f); Replay attack four signals (g-h). Replay attack five signals (i-j). Replay attack six signals (k-l). All signal values have been normalized to account for differing value ranges and facilitate visualization.

## Methodology

The proposed framework consists of three modules. The first module leverages a deep AE architecture composed of layers with time sequence learning capabilities, to form a reconstructive model of the PUR-1 reactor. The second module evaluates the reconstructive model in real time by comparing reconstructed outputs with live measurements to determine whether deviations from normal behavior are present. The third module provides local interpretability for attack type, source identification, duration and timing, through a customized SHAP-based analysis tailored for MTS data and the specifics of the AE. The complete framework is presented in Figure 9.

It is important to note that the use of an AE architecture offers key advantages. Since AEs are trained exclusively on normal operational data, they eliminate the need to obtain rare or difficult to collect anomalies. Moreover, AEs are capable of detecting previously unseen anomalies without requiring prior knowledge of the system, removing the need to manually label known irregular conditions. Consequently, no training data representing replay attacks or other abnormal scenarios are required, which is an important benefit for real-time applications in nuclear systems, where such abnormalities are scarce or even unavailable.

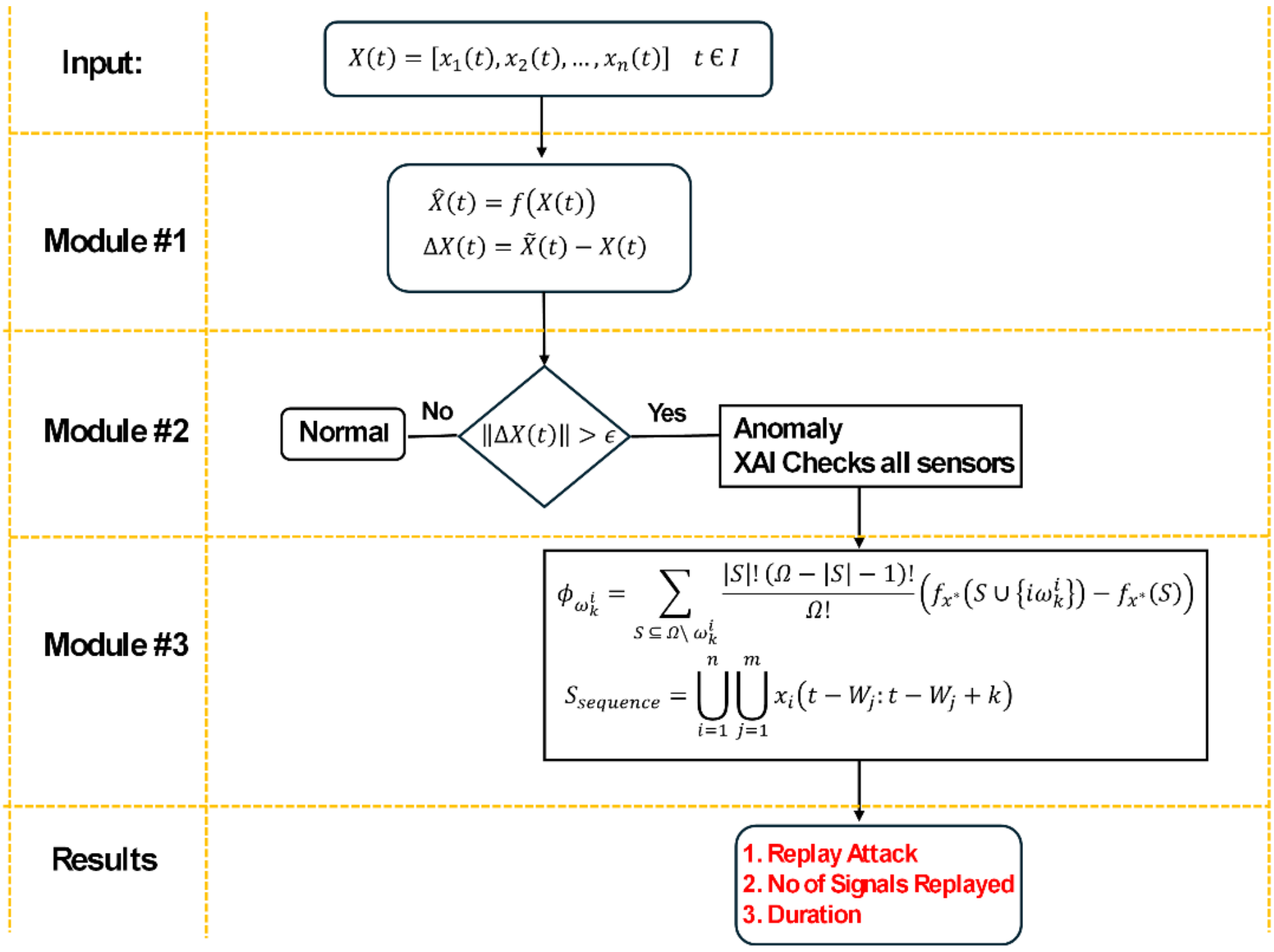

Figure 9: Methodology workflow

### Module 1: Reconstructive Modeling of Reactor Behavior

The first module involves training a deep AE architecture exclusively on normal historical data enabling the system to learn expected reactor behavior. The fundamental concept of an AE is to generate a lower dimensional representation of the input sequence through a series of neural layers and then reconstruct it back to its original dimensional space. By using a large number of full reactor power cycles (Figure 6), the model captures the inherent dynamics, correlations, and interdependencies among sensor signals across all reactor conditions (startups, steady states, transients, shutdowns). Let

$$X(t) = [x_1(t), x_2(t), \ldots, x_n(t)] \quad t \in I$$

represent a set of *n* sensor measurements over the interval *I*, where $x_i(t)$ denotes the reading from the *i-th* sensor at time $t \in I$. A reconstructive model $f$ is trained under normal operational conditions on historical data to learn the functional mapping:

$$\hat{X}(t) = f\big(X(t)\big)$$

Where $\hat{X}(t)$ is the model's reconstruction for the set $X(t)$ over the same time interval *I.* The model is optimized to minimize the error between the original and the reconstructed data for the fixed time interval *I*, over the entire length of the training data, such that:

$$\min_f \mathbb{E}\big[\big\|X(I) - \hat{X}(I)\big\|\big]$$

Where $\|\cdot\|$ represents a norm suitable for quantifying the similarity between two vectors.

A number of methods have demonstrated effectiveness in capturing temporal dependencies in MTS data, with characteristic examples being the autoregressive moving average models and their variants, correlation-based, and ensemble methods [54], [55], [56], [57], [58]. AEs have gained attention and have become the standard approach for unsupervised anomaly detection via reconstruction in time series data, due to their ability to generalize and robustness to noise [59] [60]. Although not originally designed for anomaly detection, AEs were introduced by Rumelhart et al. (1986) [61] as a novel method for dimensionality reduction and representation learning, with applications in image storage and transmission. Their application for anomaly detection became prominent after 2010 with advancements in deep learning architectures and computational capabilities. In one of the first studies, Sakurada et al., (2014) [62] showed that AEs can effectively detect anomalies through nonlinear dimensionality reduction. In [59], [63] unsupervised frameworks based on AEs for anomaly detection and diagnosis in MTS are proposed with improved performance compared to baseline methods. Despite their effectiveness, these approaches do not consider the specific patterns associated with flagged abnormalities and often rely on simplified assumptions regarding the root cause of the anomalous event. Consequently, they are not suitable when full scale interpretability is required.

AEs consist of three main components: an encoder, which compresses the input data through dimensionality reduction; a latent representation, which captures the most important patterns in the lower dimensional data; and a decoder, which reconstructs the data from the latent space back to the initial dimensions. Due to the wide range of applications, numerous AE architectures have been proposed, each tailored for a specific purpose. Most significant examples include vanilla AEs, deep AEs, convolutional AEs, RNN-based AEs, LSTM AEs and Variational AEs [64]. In our study, LSTM layers [65] are employed within the AE architecture to capture temporal dependencies and accurately reconstruct sequences. LSTM networks are a specific type of Recurrent Neural Network (RNN) [66], designed to overcome the vanishing gradient problem commonly encountered in traditional RNNs, where small gradients can impede the training process [67] [68]. LSTM networks incorporate a memory cell and a hidden state, along with input, output, and forget gates that selectively discard or retain information from the memory cell. This architecture makes LSTMs well-suited for time series data and ideal for modeling systems with time-dependent patterns, such as reactor conditions. The equations governing the LSTM cell at a timestep t are:

$$f_t = \sigma\left(W_f x_t + U_f h_{(t-1)} + b_f\right)$$
$$i_t = \sigma\left(W_i x_t + U_i h_{(t-1)} + b_i\right)$$
$$\widetilde{c_t} = \tanh\left(W_c x_t + U_c h_{(t-1)} + b_c\right)$$
$$c_t = f_t \odot c_{(t-1)} + i_t \odot \widetilde{c_t}$$
$$o_t = \sigma\left(W_o x_t + U_o h_{(t-1)} + b_o\right)$$
$$h_t = o_t \odot \tanh(c_t)$$

Where $f_t$ is the forget gate, $i_t$ is the input gate, $\widetilde{c_t}$ is the candidate cell, $c_t$ is the updated cell state, $o_t$ is the output cell, $h_t$ is the hidden state, σ is the activation function, $W_f, W_i, W_c, W_o, U_f, U_i, U_c, U_o$ are the weight matrices and $b_f$, $b_i, b_c, b_o$ are the bias terms.

**Module 2: Reconstruction Error Analysis**

The second module analyzes the reconstruction errors produced by the model in module 1 and uses them as a quantitative measure of potential anomalies in the system. Reconstruction error analysis has been successfully applied in relevant studies in the nuclear domain. The work in [69] presents an efficient AE-based methodology for predicting potential anomalies in the cooling system of a simulated PWR, using the reconstruction error between unseen and reconstructed data. Similar studies include [70], [71], [72]. The main difference in our study is that the anomaly does not appear as an obvious temporal trend, but rather as a disruption in the expected intercorrelation among signals under a dynamic operating condition, which is ultimately driven by reactor operation. In this context, the replay attack introduces a conceptual inconsistency in the data [41], that only a trained operator with knowledge over the system would recognize. This distinction is important because it indicates that such anomalies would remain undetected by methods relying solely on temporal patterns or traditional detection approaches.

To identify a replay attack, a metric that triggers a flag at a given time is required. Let's assume a modified measurement vector:

$$\tilde{X}(t) = [\, \tilde{x}_1(t), \tilde{x}_2(t), \dots, \tilde{x}_n(t) \,]$$

Where $\tilde{x}_i(t)$ represents the replayed value of the *i-th* sensor at time *t*. Since the model is trained solely on normal data, it is expected to reconstruct any unseen conditions poorly. The deviation introduced by the replay attack can be calculated as:

$$\Delta X(t) = \tilde{X}(t) - X(t), \ \ t \in I$$

Where $\tilde{X}(t)$ is the reconstructed data of $X(t)$ at time $t$. An anomaly is flagged for a potential replay attack if:

$$||\Delta X(t)|| > \epsilon, \qquad for\ t \in I$$

Where $I$ is a fixed time interval within the time series, and $\epsilon$ is the minimum deviation required to trigger the system's response. The threshold $\epsilon$ defines the sensitivity of the system and was evaluated as part of the hyperparameter tuning. Module 2 continuously evaluates the residual errors over a predefined window $I$ across the entire time series. Sections of the time series with errors below the threshold are considered normal reactor operation. For sequences exceeding the threshold, a post hoc interpretability analysis is employed to determine whether patterns exist that justify the presence of a replay attack in the system.

**Module 3: Source and Timing Identification with Post Hoc Explainability**

Detecting a deviation alone is not sufficient to confirm a replay attack, nor does it reveal the source of the anomaly, since deviations may arise from routine operational changes, measurement noise, or even sensor malfunctions. This module introduces a modified SHAP-based analysis which, combined with the results from modules 1 and 2, provides an additional layer of explainability, enabling the identification of falsified sensors, the timing and duration of the attack.

While AEs and other AI/ML algorithms have been successfully applied across a broad range of engineering and industrial domains, their limited explainability, especially in the context of multi-signal monitoring and human-scale reasoning, complicates their adaption in critical infrastructure [73]. Prior research [74] [75] has shown that lack of trust in advanced data analytics can be attributed to the difficulty of gaining clear insights from such models. To address this limitation, SHAP analysis has recently emerged from game theory as a model-agnostic framework for explaining ML predictions [76] [77].

The SHAP framework quantifies each feature's contribution to a model's output for a specific input by assigning Shapley values that either support or offset the prediction. The Shapley value for a feature $i$ at a given input $x = x^*$ s defined as:

$$\phi_i = \sum_{S \subseteq N \{i\}} \frac{|S|!\,(p - |S| - 1)!}{p!} (f_{x^*}(S \cup \{i\}) - f_{x^*}(S))$$

Where, $\phi_i$ denotes the Shapley value calculated for feature $i$, N is the set of all features, p is the total number of features in the dataset, $S$ represents any subset (or coalition) of features not containing $i$, |S| is the number of features in subset $S$, and $f(S \cup \{i\} - f(S)$ is the marginal contribution of feature $i$ to the coalition S. The summation runs over all possible subsets $S$ of features, excluding feature $i$. The function *f(S)* represents the model output when only the features in subset S are known, while the remaining features are marginalized out. It is defined as:

$$f(S) = E[f(X) \mid X_S]$$

Where *f(X)* is the model's prediction over all features, and the expectation is taken over the marginal distribution of the features not included in subset S. This operation averages the model output across all possible values of the missing features. The term *E[f(X)]* represents the expected model output over the full feature space and serves as the baseline reference. The sum of all Shapley values is equal to the difference between the model's output and the expected output at the baseline (background) *E(f(X))*. Each individual value $\phi_i$ represents the contribution of feature *i* to that difference.

# Implementation

### Data processing

The datasets used for training were preprocessed to ensure compatibility with the AE model. Data was collected from the monitoring software of the digital twin system (Figure 3), either in real time during reactor operation, or at a later time by selecting specific operational periods. Each dataset was normalized using a Min-Max scaler to the range between 0 and 1. To maintain consistent scaling across all datasets, the minimum and maximum values for each feature were derived from the last three years of reactor operation. Each dataset, corresponding to a distinct reactor operation, is converted on the fly into 3D tensors and fed to the AE model during training. The tensors are constructed by sliding a one-second window across the dataset, as shown in Figure 10. This process generates tensors with dimensions (batch size x window length x features). This format is well-suited for MTS data, ensuring that during testing the model consistently considers the past window of seconds at each time step and reconstructs the same window length for further analysis. The optimal window length was found to be 10 seconds.

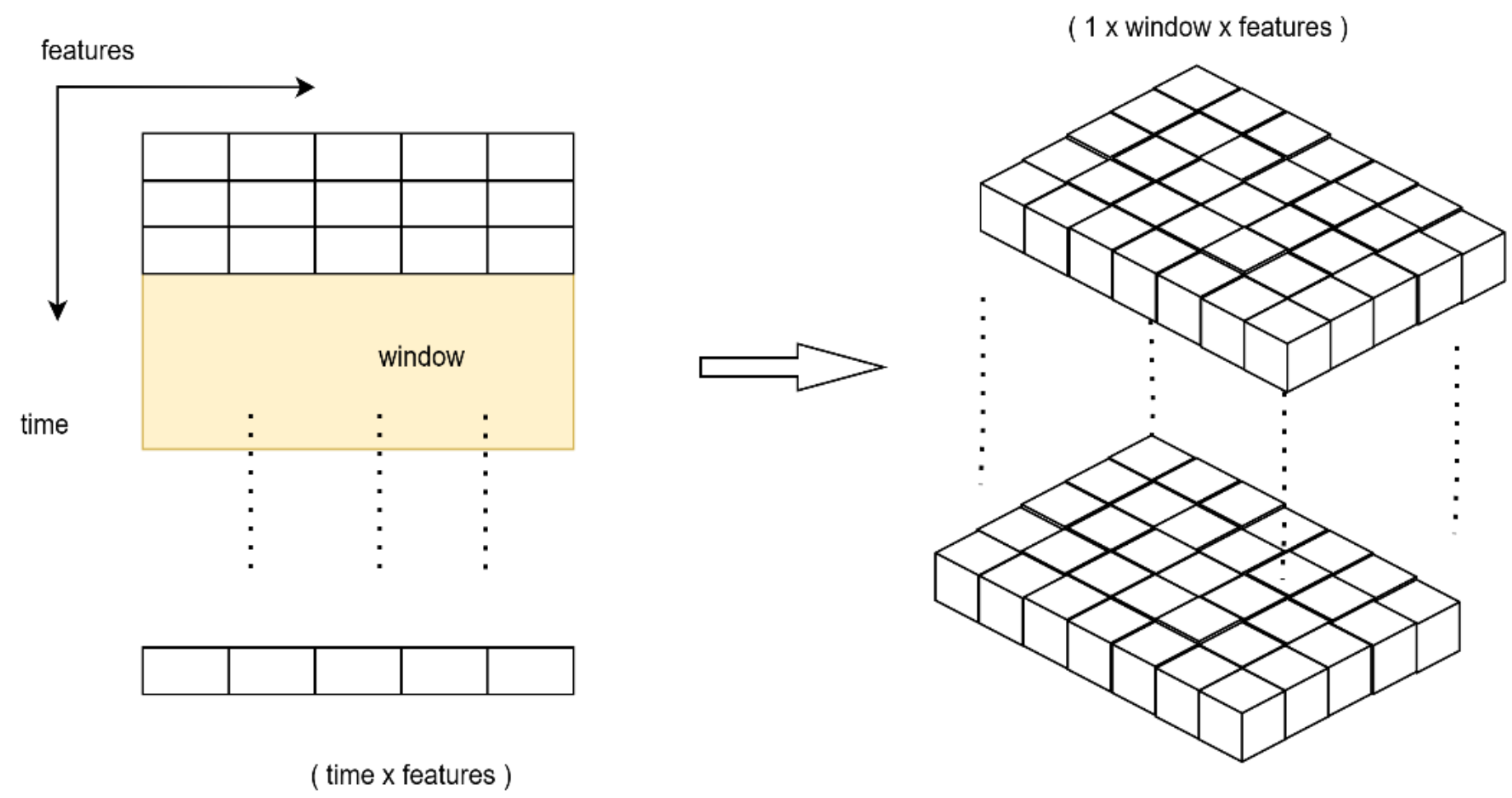

Figure 10: Tensors are created by sliding a one-second window across the dataset. This process generates (dataset length-window) tensors, each with dimensions (1 x window size x features).

**Training, tuning, implementation**

The AE model was initially trained on the full cycle data (Figure 6) to capture the reactor dynamics, followed by transfer learning on the SCRAM datasets (Figure 7) to refine its ability to detect extreme signal variations. To optimize the AE architecture, hyperparameter tuning was conducted using Bayesian optimization [78]. The tuning step is essential in any ML application, as hyperparameter selection highly affects model performance [79]. Bayesian optimization was selected over other common methods (e.g., grid search, random search, etc.) due to the noisy nature of the data and the complexity of the deep architecture employed.

The AE architecture is designed to be flexible in hyperparameter tuning. Tunable parameters include the depth and width of the LSTM layers across the encoder, decoder, and latent representation. The search space includes a range of values for the number of hidden layers, units per layer, batch size, dropout regularization to reduce overfitting, and the learning rate. Of the 47 full reactor cycle datasets, 34 were used for training during hyperparameter optimization, 10 for validation, and the remaining 3 were used for testing after the tuning was completed. The tuner evaluated 20 different hyperparameter combinations from the predefined search space, with each model trained for 30 epochs. The optimal combination was selected based on minimizing the Mean Squared Error (MSE) on the validation test. Table 3 lists the optimal hyperparameters found, and Figure 11 presents the corresponding AE architecture.

After identifying the optimal architecture, the network was retrained on the SCRAM datasets to obtain the final reconstructive model. Of the 24 available datasets, 20 were used for training and 4 for the validation. Figure 12 shows the training and validation loss during training. The number of training epochs was determined based on the convergence observed in preliminary experiments. The model was initially trained for 60 epochs (Figure 12), during which both the training and validation losses exhibited no further improvement after approximately 25 epochs. To mitigate potential overfitting and reduce computational cost, the final training was limited to 30 epochs.

Table 3: Optimal hyperparameters for the AE model determined through tuning

| **Hyperparameter** | |
|---|---|
| Learning Rate | 0.000352 |
| Encoder Hidden Layer #1 | 256 neurons |
| Dropout Layer | 0.1 |
| Encoder Hidden Layer #2 | 128 neurons |
| bottleneck | 32 neurons |
| Decoder Hidden Layer #1 | 128 neurons |
| Dropout Layer | 0.2 |
| Decoder Hidden Layer #2 | 128 neurons |
| Batch size | 32 |
| Window size | 10 |

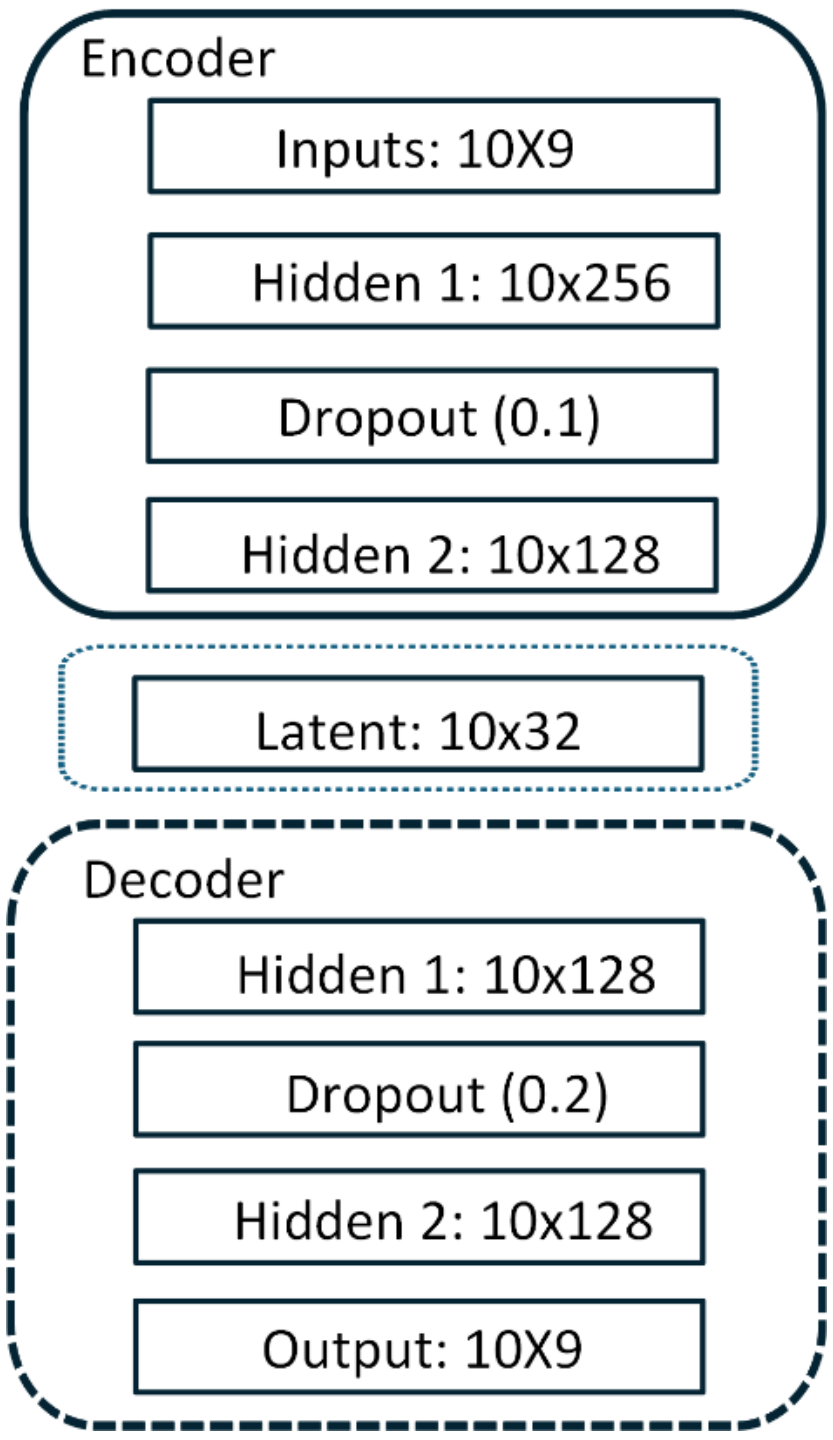

Figure 11: Architecture of the AE model.

A Python package named Nuclear-replay-attack-detection accompanies this paper and is publicly available at https://github.com/Kvasili/Nuclear-replay-attack-detection. The package includes an example of the proposed framework on the data used to simulate the replay attacks. Model training, testing, and tuning were performed using the Tensorflow, Keras, scikit-learn, and Pandas libraries. All simulations were executed on an NVIDIA GeForce GTX 1650 Ti GPU, with a training time of approximately 4 hours for 30 epochs on the full cycle datasets.

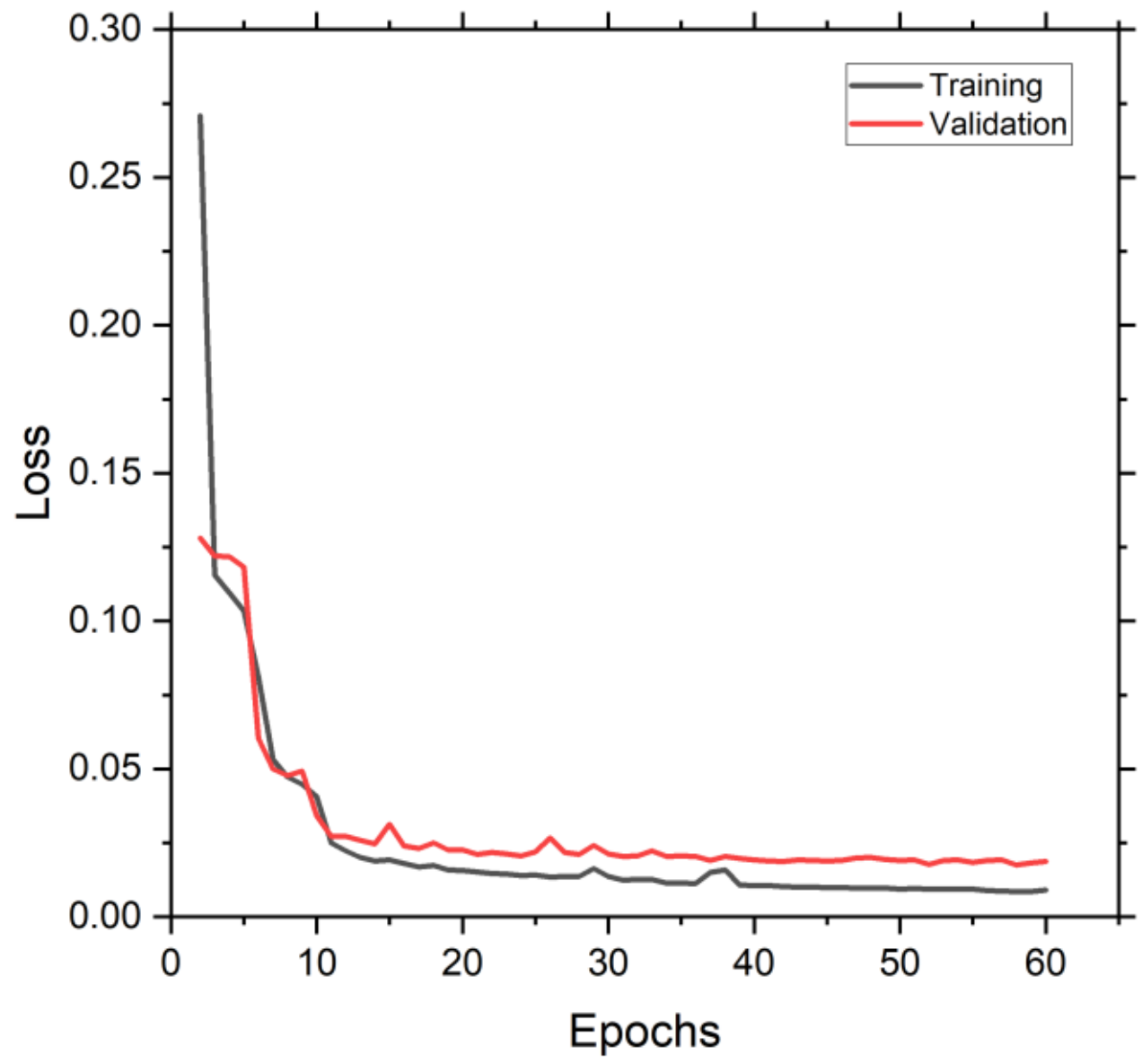

Figure 12: Training and validation losses for the optimized architecture over 60 training epochs.

To evaluate the model's performance on unseen data, the trained model was applied to the three testing datasets of the full cycle reactor operation. Figure 13 shows a comparison between the reconstructed and original data for the neutron counts feature (ch.1). The model successfully reconstructs the signal in all three cases, proving that the reconstructive modeling of reactor behavior is effective. During steady state operation, the reconstruction error remains consistently low, whereas spikes in error occur during changes in reactor conditions, particularly during startup and shutdown when extreme transients take place. During these events, it is expected that a falsified signal will produce even higher disagreement between the model and the actual behavior and will allow the identification of potential anomalies through module 2 and 3.

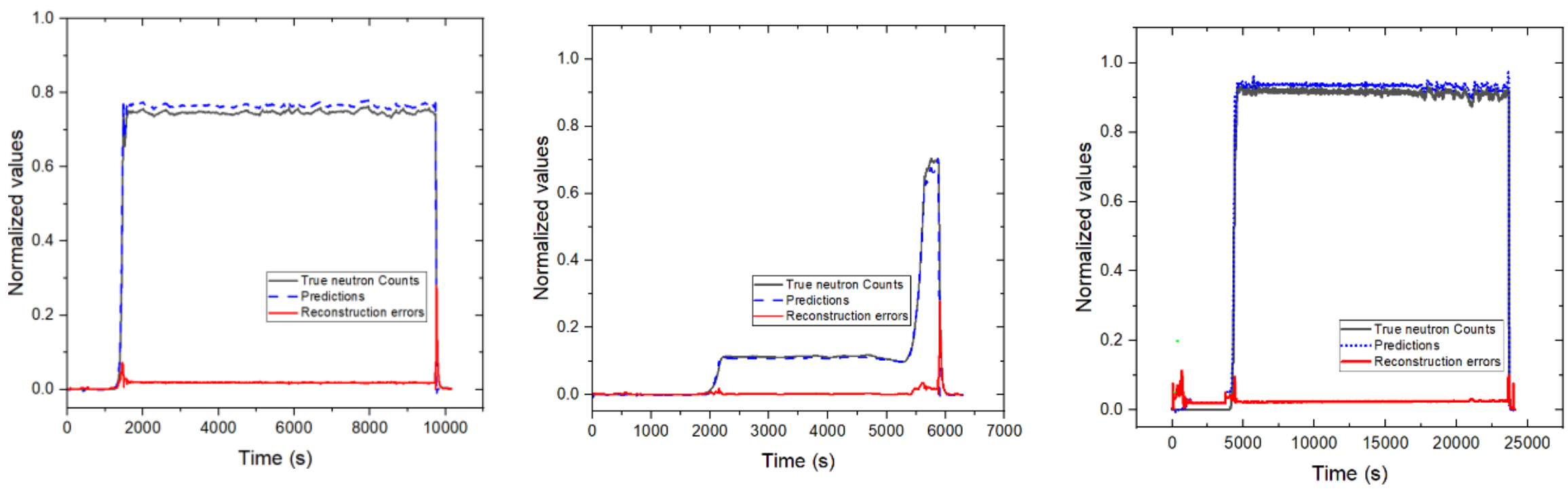

Figure 13: AE model reconstructions and errors on the testing datasets.

**Reconstruction error analysis**

The model's loss objective is to minimize the reconstruction errors of sequences representing normal operation during training, so that normal features are accurately reconstructed while erroneous features are reconstructed poorly during testing. Anomalies can be detected by evaluating the magnitude of either the Mean Squared Errors (MSE) or the Mean Absolute Errors (MAE), given by the following formulas:

$$MSE = \frac{1}{N} \sum (X - X')^2$$

$$MAE = \frac{1}{N} \sum | X - X' |$$

Where N is the number of samples in the training set, corresponding to the number of tensors created. $X$ is the vector of sensor values within the predefined window, and $X'$ is the equivalent vector with the predicted values. Following the approach commonly adopted in related studies [69] [71], the anomaly threshold is determined from the distribution of MSE or MAE on the normal data. By examining this distribution, one can assess how well the model reconstructs normal sequences and then define a threshold such that any sequence whose reconstruction error exceeds this value is classified as an abnormal pattern.

Figures 14.a and 14.b show the reconstruction error distribution, calculated using the MAE over the windowed sequences of the training and validation datasets. To evaluate the performance, Figure 14.c presents the error distribution in one of the falsified datasets used to simulate a replay attack. The visualization reveals nearly identical left-skewed distributions for both the training and validation datasets, with the vast majority of the reconstructed sequences concentrated near the zero-error bin. This indicates that the model is effective at reconstructing normal behavior. Larger error values, extending beyond 0.2, are also observed but with a significantly lower frequency. These higher errors are likely attributed to inherent system noise or sensor-related artifacts. The pattern differs in Figure 14.c. While a large concentration of small error values remains, which fall into the statistical uncertainty of the normal datasets, a substantial additional concentration of high error values appears around the 0.4 bin. This indicates that a portion of the reconstructed sequences clearly deviates from the expected behavior, suggesting the presence of a potential anomaly in the system.

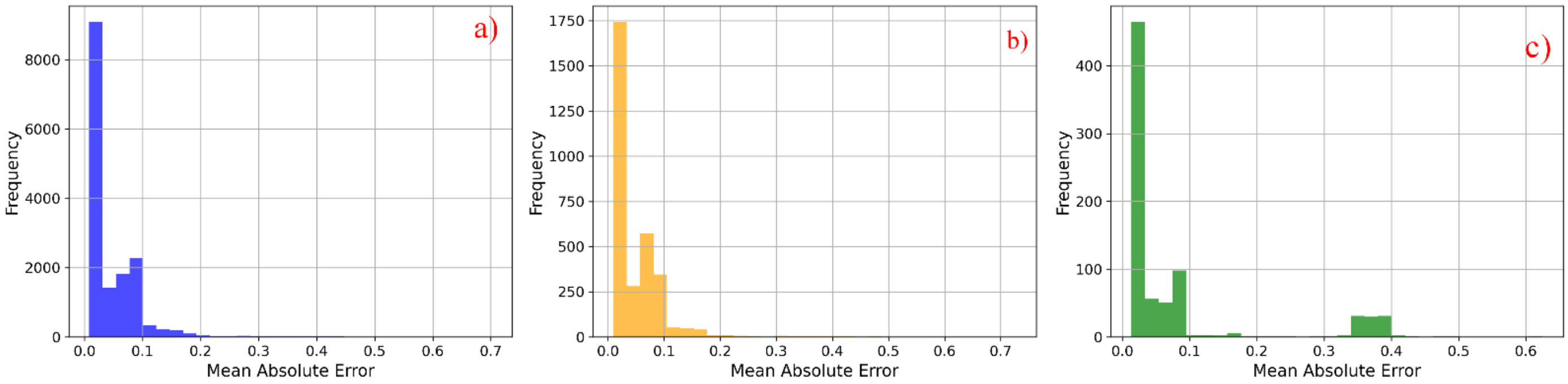

Figure 14: Reconstruction error distributions for training (a), validation (b), and testing (c) datasets.

**XAI implementation**

Although SHAP was originally introduced to provide global explanations, the increasing complexity of the models shifted the focus toward instance-level explanations [42]. In this work, we further emphasize the importance of time-sequence-level explanations, particularly in cases where the underlying dynamics change over time.

Applying the original SHAP framework directly to MTS data becomes prohibitively expensive, because the computation of Shapley values scales exponentially with the number of features. In an MTS setting, each feature at each timestep would be treated as a separate input dimension, requiring the computation of Shapley values over all possible coalitions of feature-time pairs. This would result in a combinatorial explosion that makes the direct application of standard SHAP computationally infeasible. Moreover, even if computational cost was not a limitation, this approach could mask meaningful temporal patterns by spreading feature contributions across a large number of timesteps. To overcome this, a windowed approach, called windowSHAP, was proposed in this study [80] for application to time series data. In this method, the time series is divided into non overlapping windows, and each window is treated as a single input unit for the SHAP calculation. This restricts the computation to distinct temporal segments rather than all possible combinations of feature-time points, significantly reducing the complexity while preserving interpretability in both the temporal and feature dimensions. Considering each window as an input feature to windowSHAP, the Shapley value $\phi_{\omega_k^i}$ for the k$^{th}$ window $\omega_k^i$ is calculated as:

$$\phi_{\omega_k^i} = \sum_{S \subseteq \Omega \setminus \omega_k^i} \frac{|S|!\,(\Omega - |S| - 1)!}{\Omega!} \left(f_{x^*}\left(S \cup \{i\omega_k^i\}\right) - f_{x^*}(S)\right)$$

Here, the notation follows the standard SHAP formulation, with the key difference that the features correspond to time windows instead of the p input variables. Ω denotes the set of all time windows for all variables.

For our application, windowSHAP was modified to account for the specifics of the AE architecture and the dynamics of the critical event in the reactor system. WindowSHAP distributes Shapley values based on ML predictions, however this is not directly applicable to an AE model, which reconstructs the full input sequence than producing a single scalar prediction. Instead of applying windowSHAP directly to the entire MTS, which raises concerns for real-time detection, we applied the modified version in real time only to the reconstructed sequences that are flagged as anomalous. The flagged sequence is divided into windows, and the Shapley values for the full sequence are calculated as:

$$S_{sequence} = \bigcup_{i=1}^{n} \bigcup_{j=1}^{m} x_i\left(t - W_j : t - W_j + k\right)$$

Where $i$ indexes the features, $j$ indexes the windows within the flagged sequence, $\cup$ denotes the union over all features and windows, $x_i$ is the $i$-th feature, $W_j$ is the offset for the $j$-th window, $k$ is the window size, and $x_i(t - W_j : t - W_j + k)$ is a subsequence of values of feature $i$ within the $j$-th window. In practice, this formula represents the partitioning of the sequence into distinct windows, over which the modified windowSHAP calculates Shapley values to identify the timesteps and features within the sequence that have the most influence on the model's reconstruction error.

In addition, windowSHAP occludes data by replacing the "missing" values with the global average. This is also not suitable for our case because the occluded values represent a baseline and expected behavior of the system. In contrast, a forced shutdown event (reactor SCRAM) can be initiated at any time during reactor operation, often when the reactor is operating at a power level sufficiently far from the mean. Therefore, substituting a sudden change in power with the global mean would not represent the expected or physical realistic baseline for reactor behavior in this context. To accurately represent the baseline condition during a SCRAM event, the baseline was set as the expected signal behavior for all signals in the dataset during the event (Figure 15). Since the dynamics of the signals during the SCRAM event are known, the baseline is dynamically updated over time to account for signal variability as the event evolves. In this way, the algorithm assigns importance to the features whose evolution within the input time window produces the largest change in the reconstruction error.

The interpretation of Shapley values depends directly on the choice of the baseline data. Negative Shapley values indicate that deviations in a certain feature decrease the error, suggesting a correlation with the actual signal. Positive Shapley values indicate that deviations in an input feature increase the error, suggesting a lack of correlation with the actual signal. Shapley values near zero indicate that the signal is at the baseline or that the variation from the baseline did not influence the error [33].

By applying the modified WindowSHAP method, a characteristic pattern emerges in the Shapley value distribution of each feature during a replay attack. This pattern provides an interpretable indicator that helps distinguish replay attacks from other abnormal conditions, such as FDI or sensor malfunctions. Compared to distance-profile-matrix approaches used in related work, where each flagged sequence is compared against all sequences from prior timesteps, this method is less computationally expensive, while simultaneously accounts for multiple replayed signals.

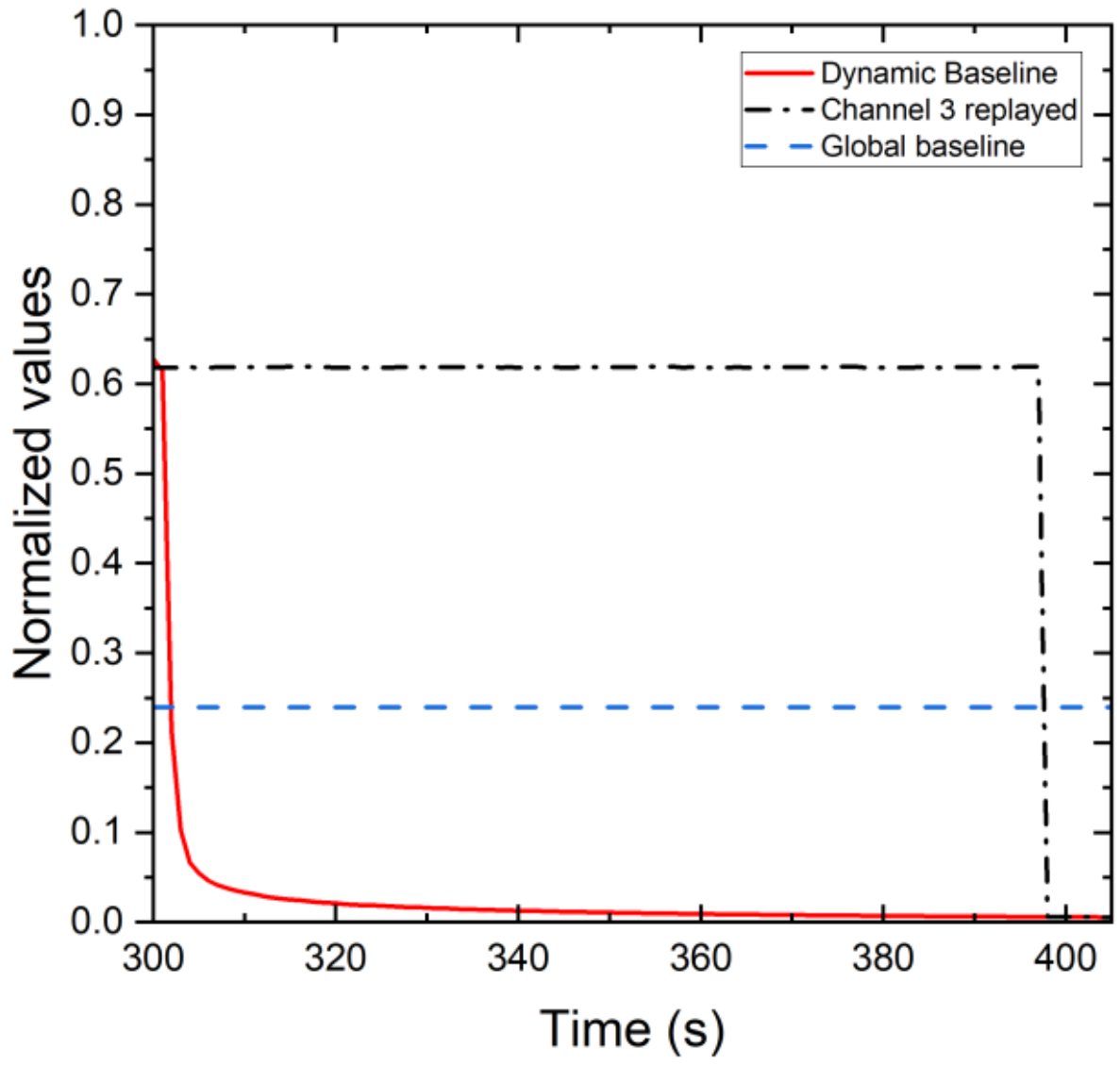

Figure 15: Comparison between the dynamic baseline which represents the expected signal behavior (red line), and the global mean baseline (blue line) for reactor power during the SCRAM event. The expected behavior provides a physically meaningful dynamic baseline for the windowSHAP occlusion, whereas the global mean does not capture the transient dynamics.

# Results

To detect anomalies effectively, an appropriate threshold must first be determined to indicate deviations from normal patterns (module 2). The reconstruction error is evaluated per second, using a one-second sliding window which considers the preceding 10 seconds. For each time step, the model computes the reconstruction error over this window, and an anomaly is flagged whenever the error exceeds the defined threshold. By comparing the flagged anomalies with the true labels, an anomaly detection metric is calculated. Table 4 presents the anomaly detection performance for error thresholds ranging from 0.05 to 0.25 across all datasets. An error threshold between 0.1 and 0.25 results in accuracies between 93% and 98% for the normal dataset. A threshold of 0.15 achieves a detection accuracy above 96% for all datasets. Identifying anomalies through a simple threshold is the most straightforward form of reconstruction-based analysis and is widely adopted in related studies.

Table 4: Detection Accuracy for Various Thresholds

| Threshold | Normal | Replay#1 | Replay#2 | Replay#3 | Replay#4 | Replay#5 | Replay#6 |
|---|---|---|---|---|---|---|---|
| 0.05 | 0.65 | 0.77 | 0.77 | 0.77 | 0.77 | 0.77 | 0.77 |
| 0.1 | 0.93 | 0.97 | 0.97 | 0.97 | 0.97 | 0.97 | 0.97 |
| 0.11 | 0.94 | 0.97 | 0.97 | 0.97 | 0.97 | 0.97 | 0.97 |
| 0.12 | 0.95 | 0.96 | 0.97 | 0.97 | 0.97 | 0.97 | 0.97 |
| 0.13 | 0.95 | 0.96 | 0.97 | 0.97 | 0.97 | 0.97 | 0.97 |
| 0.14 | 0.96 | 0.96 | 0.97 | 0.97 | 0.97 | 0.97 | 0.97 |
| 0.15 | 0.97 | 0.96 | 0.97 | 0.97 | 0.97 | 0.98 | 0.98 |
| 0.2 | 0.98 | 0.88 | 0.97 | 0.98 | 0.97 | 0.98 | 0.98 |
| 0.25 | 0.98 | 0.86 | 0.91 | 0.98 | 0.98 | 0.98 | 0.98 |

The XAI module employs the modified windowSHAP algorithm to quantify the contribution of each feature at each time step to the reconstruction error of a flagged window. Specifically, each Shapley value represents the contribution of a signal segment to the difference between the reconstruction error of the flagged window and the error obtained when that segment is replaced by its baseline (expected) behavior. A negative Shapley value indicates that this portion of the signal is consistent with the expected baseline behavior (Figure 15), since it reduces the reconstruction error. Conversely, a positive Shapley value indicates that the portion of the signal deviates from the baseline and increases the error. Values near zero indicate close agreement with the baseline. After computing the Shapley values for each flagged 10-second window, the values are averaged across the window to yield a single representative value per second for each signal.

A consistent pattern emerges when analyzing the distribution of Shapley values across the replayed sequences. Figure 16 presents the Shapley values at the time steps where the reconstruction error exceeds the threshold, which correctly occurs during the falsification interval (300-400 s) in all six cases. For the three signals corresponding to neutron counts (ch.1), reactor power (ch.3), and neutron flux (ch.4), the Shapley values begin near zero and increase approximately linearly over the first 15 seconds. This trend reflects the time required for the true signals to decrease to low values during the event. The increasing Shapley values therefore indicate a progressively increasing disagreement between the expected behavior, which should exhibit a rapid decline, and the replayed behavior, which instead repeats steady state patterns. For the next 80 seconds, the Shapley values exhibit a steady, repeating pattern that reflects a sustained level of disagreement between the replayed and the expected signals, which have by this time stabilized at low values. Finally, when the attack ends at 100 seconds, the Shapley values rapidly decrease toward zero, indicating that the disagreement diminishes as the signals return to their expected values. This characteristic pattern appears consistently across all cases, indicating the same type of attack, with the number of affected signals increasing in each case.

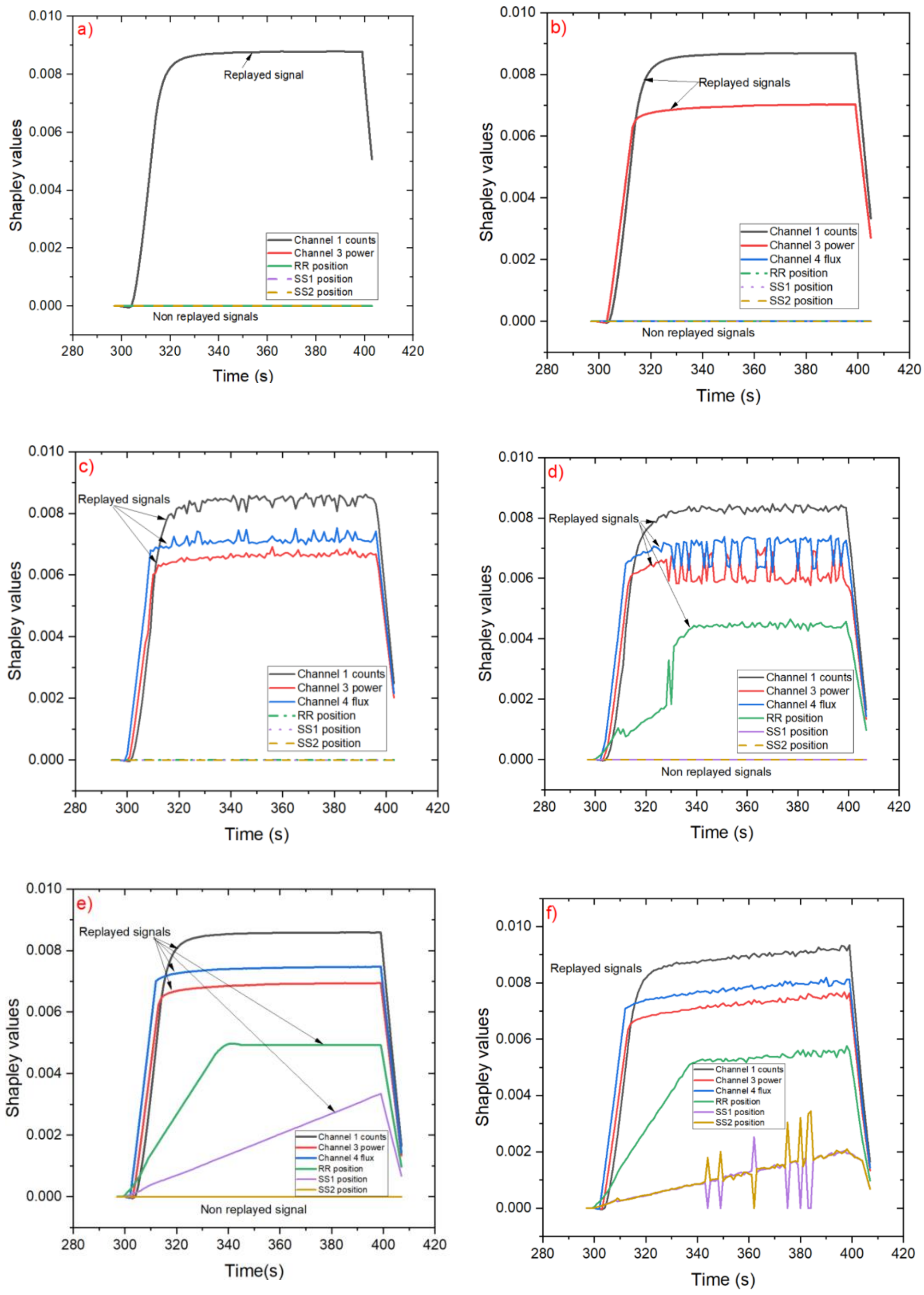

Figure 16: Distribution of Shapley values per feature throughout the attack period for a) Replay#1, b) Replay#2, c) Replay#3, d) Replay#4, e) Replay#5, f) Replay#6 cases.

For the three signals measuring the control rod positions (RR, SS1, and SS2), a similar pattern is observed. The RR rod (green line in Figure 16d-e-f) is rapidly inserted into the core during the SCRAM event, while the falsified signal remains steady throughout the entire duration. The Shapley values exhibit a nearly linear increase in disagreement for the first 30 seconds, corresponding to the time required for the RR rod drives to complete their insertion. Thereafter, the Shapley values remain approximately constant, indicating a persistent disagreement between the actual signal, which is now fully inserted, and the high values in the replayed signal. In the final seconds, the Shapley values drop quickly and approximately linearly as the signal returns to its normal behavior once the attack ends. For the SS1 and SS2 rods, the Shapley values (purple and yellow lines in Figure 16e-f respectively) exhibit a nearly linear increase in disagreement for most of the attack duration. This behavior corresponds to the steady replayed signals, while the actual signals gradually decrease until a few seconds before the attack ends, at which point the Shapley values drop sharply. Although the overall trend is linear, some spikes are observed in the SS1 and SS2 Shapley values in the last case (Figure 16f), which can be attributed to outliers caused by signal noise.

As observed, increasing the number of falsified signals results in smaller Shapley values for each additional signal at every time step. This occurs because the reconstruction error difference is distributed across a larger number of signals, causing the most influential to dominate the overall deviation between the falsified signals and the baseline. The repeated patterns appearing in the extracted Shapley values for all falsified signals directly result from the repetitive structure of the replay attack. In contrast, a different type of falsification (e.g. an FDI attack) or a falsification during a steady state reactor operation, would be expected to produce different Shapley value evolution over time, reflecting different underlying dynamics. This distinction supports the conclusion that the detected intrusion corresponds specifically to a replay attack. A research direction would be to investigate how the temporal and feature distribution of Shapley values varies with different type of falsification scenarios.

Figures 17-22 present the application of the proposed framework. The signals are first normalized and processed in real time. For each time step, the reconstruction error over the preceding 10-second window is calculated by the first and second modules. If the error falls below the threshold, the corresponding window is marked in green, otherwise it is marked in red. When an abnormal window is detected, the XAI module is activated to analyze the 10-second segment and identify any falsified signals, highlighting them in red. Once the replay attack concludes, the framework reveals the timing, duration, and number of signals that were falsified.

Table 5 shows that more than 95% of the duration of each replayed signal is correctly identified across all cases, with only minor misclassifications occurring near the beginning and end of the attack. Frame (f) in each figure displays the final set of detected replayed signals, accurately ranging from one to six. In frame (c), although the framework correctly flags an anomaly, the SHAP analysis does not yet reveal disagreement with the expected behavior. This occurs at the start of the SCRAM process, when the signals have not yet diverged significantly to form a clear pattern, normal or abnormal. After the first 10 seconds, once patterns emerge, the XAI module consistently identifies the replayed signals in all cases.

Additionally, in frame (f) the falsified signals appear to persist for approximately 10 seconds beyond the actual end of the attack. This is a direct consequence of the 10-second sliding window, in which the average Shapley value over the window is used to characterize each second. Therefore, even if only one large value exists while subsequent values drop to zero, the window may still be flagged as abnormal. This effect rapidly disappears once the entire window contains no high Shapley values, as shown immediately after the attack concludes in frame (f).

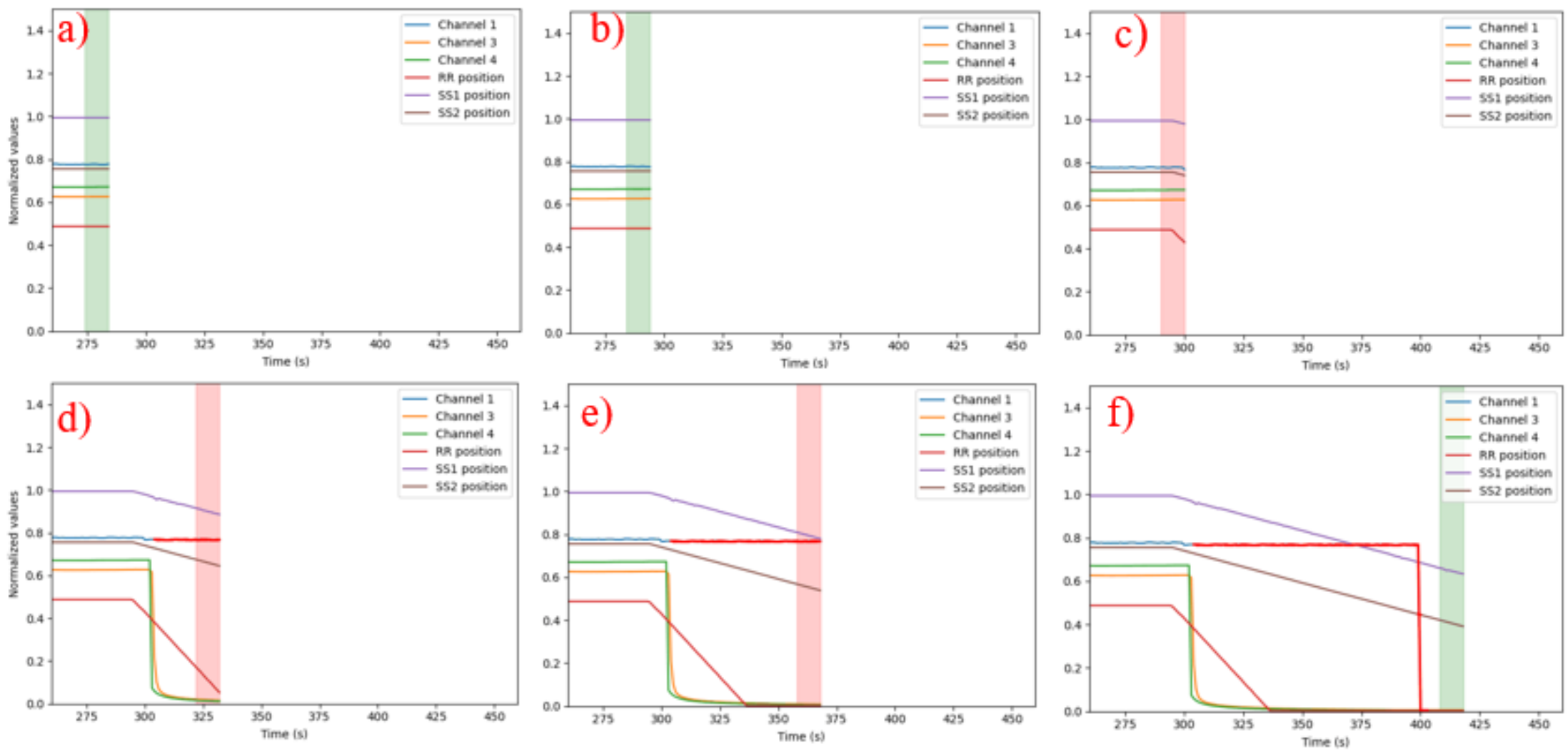

Figure 17: Replay attack detection for the Replay#1 dataset.

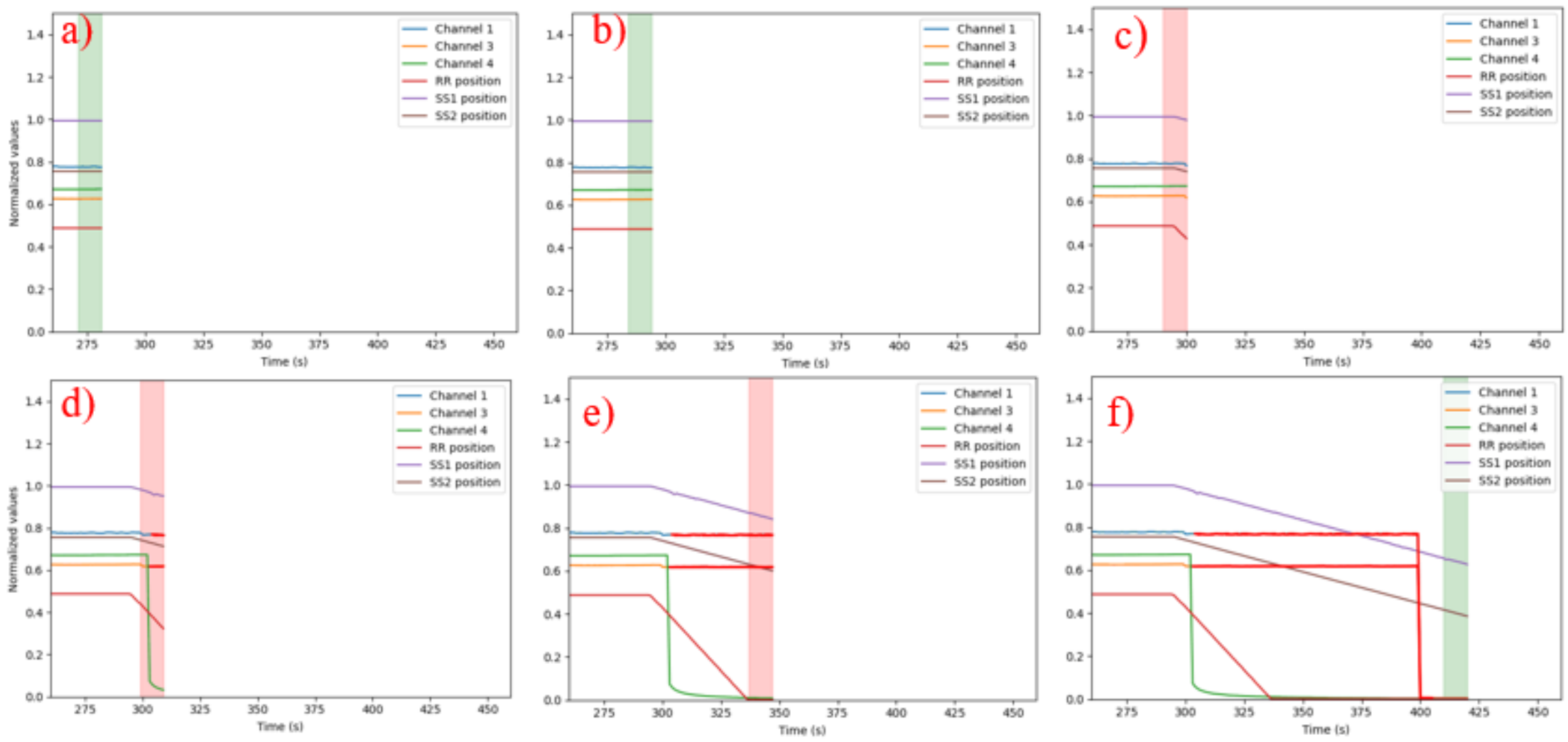

Figure 18: Replay attack detection for the Replay#2 dataset.

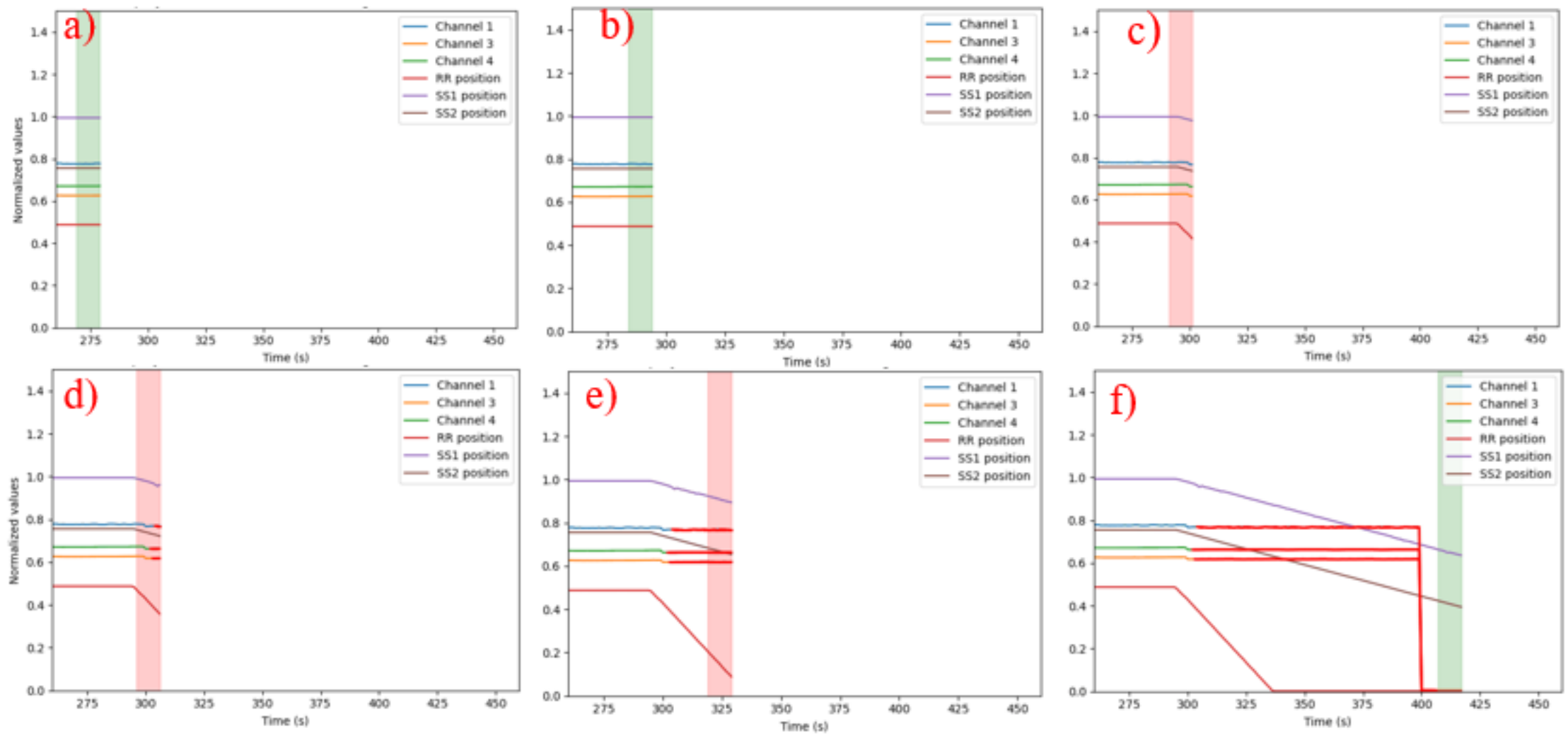

Figure 19: Replay attack detection for the Replay#3 dataset.

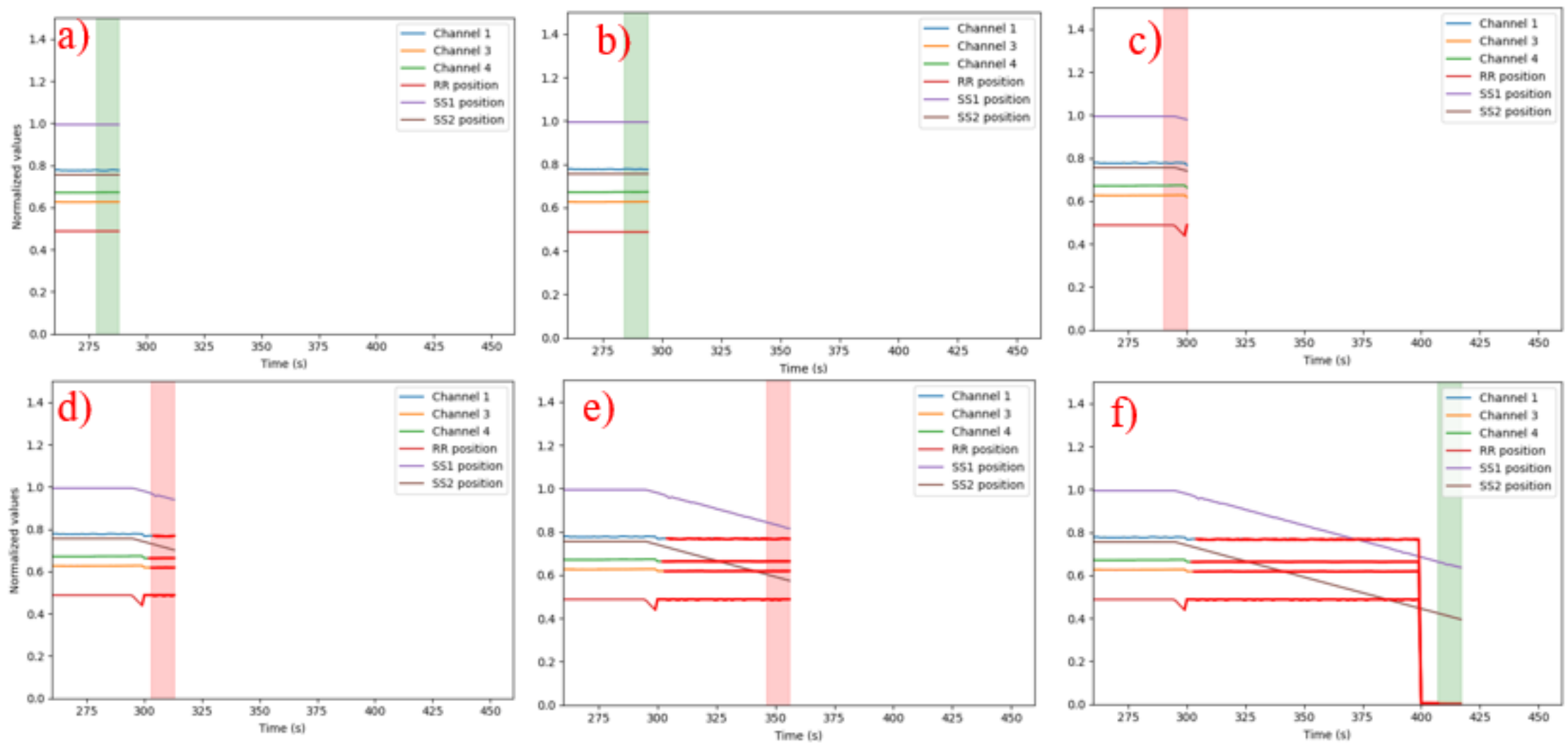

Figure 20: Replay attack detection for the Replay#4 dataset.

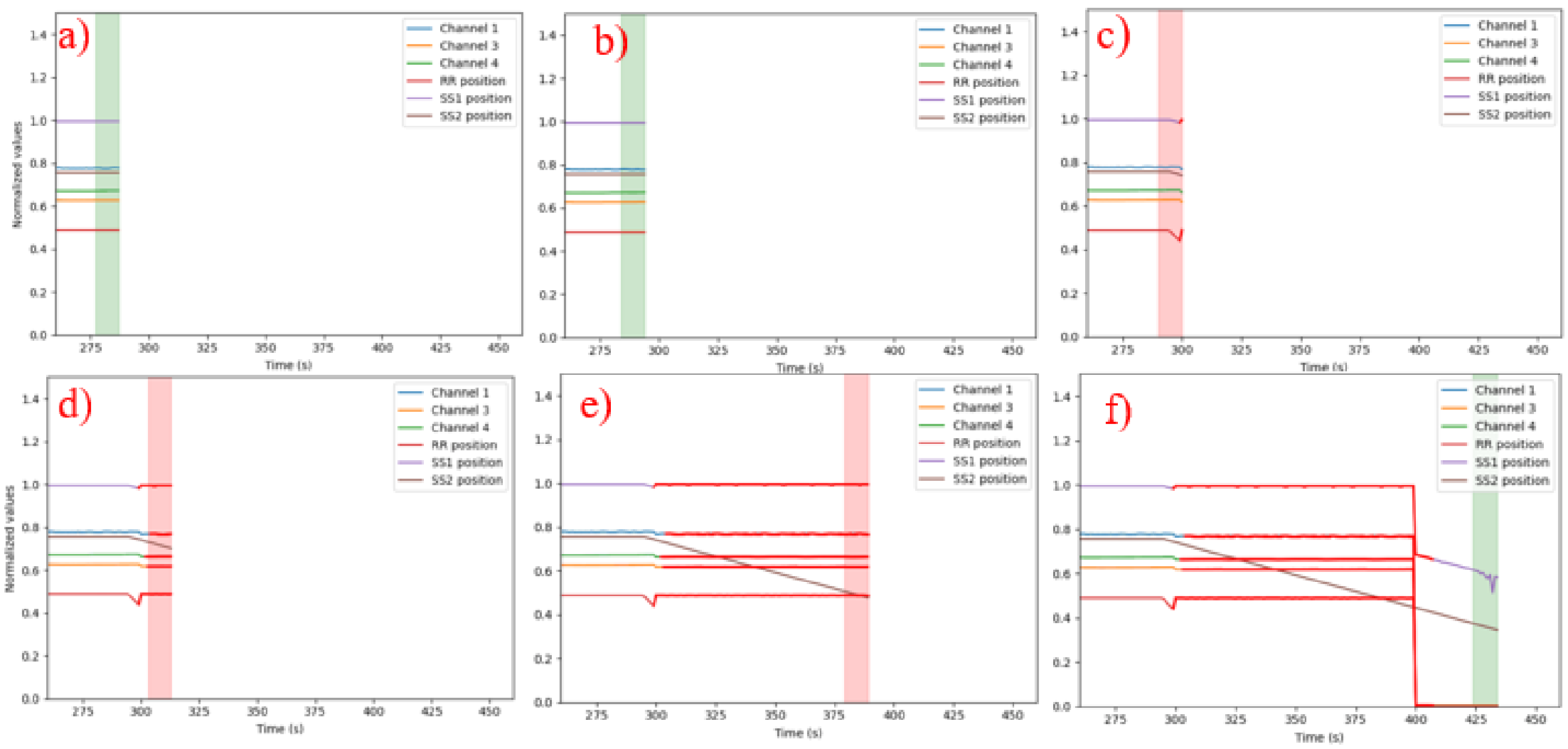

Figure 21: Replay attack detection for the Replay#5 dataset.

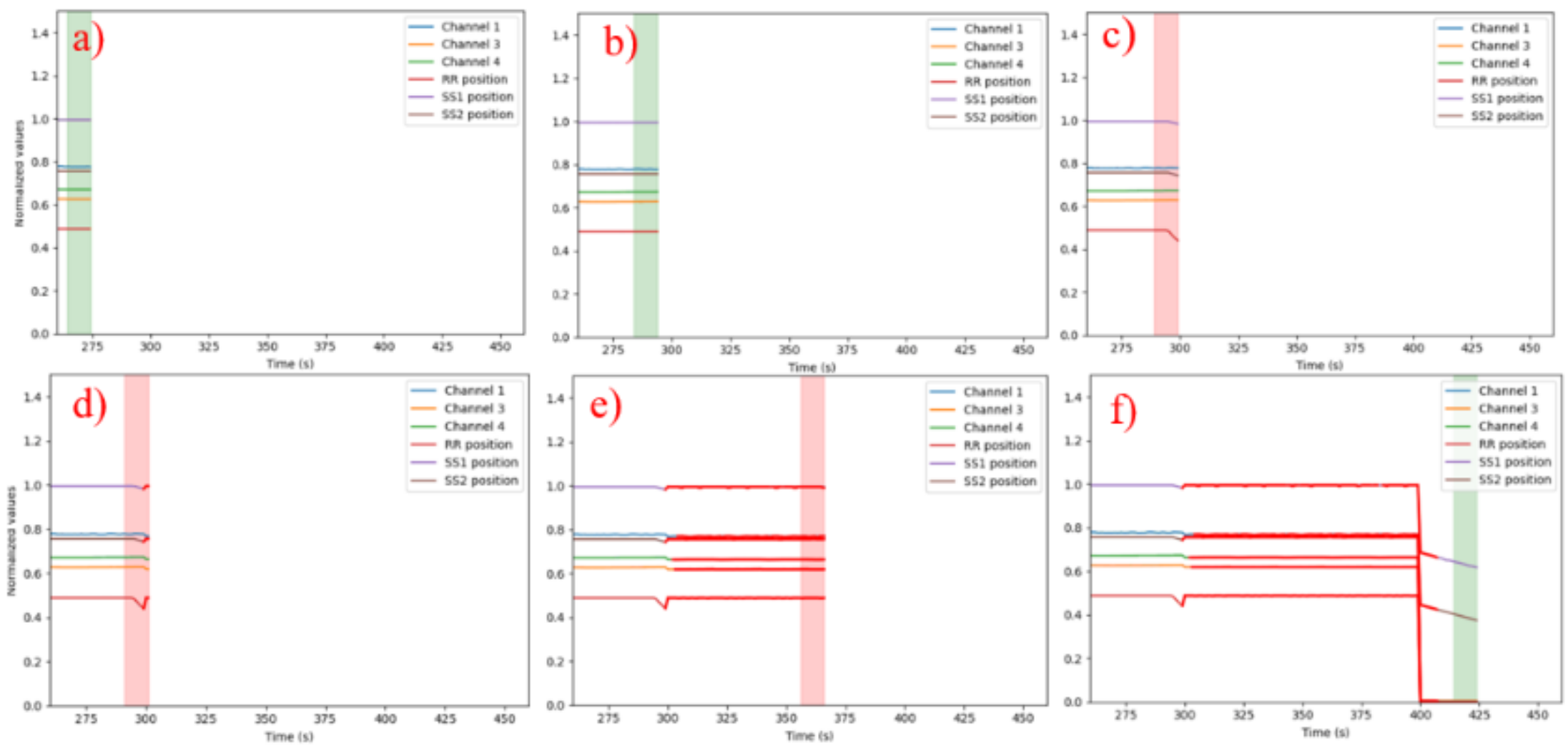

Figure 22: Replay attack detection for the Replay#6 dataset.

Table 5: Percentage of each signal accurately identified as replayed per dataset

| | Accuracy | | | | | |
|---|---|---|---|---|---|---|
| | Replay#1 | Replay#2 | Replay#3 | Replay#4 | Replay#5 | Replay#6 |
| Channel 1 | 95% | 95% | 95% | 95% | 95% | 95% |
| Channel 3 | | 96% | 96% | 96% | 96% | 96% |
| Channel 4 | | | 97% | 97% | 97% | 97% |
| RR | | | | 99% | 99% | 99% |
| SS1 | | | | | 99% | 99% |
| SS2 | | | | | | 99% |

## Conclusions

In this work, we proposed a replay attack detection and localization framework that integrates an optimized deep AE architecture for nuclear MTS data with a modified XAI module based on the windowSHAP algorithm. The network is trained solely on normal reactor data, eliminating the need for manual labeling or predefined abnormal classes. We examined a dynamic and time evolving use case of reactor operation to simulate the replay attacks, with an increasing number of concurrently falsified signals ranging from one up to six. The framework successfully triggers the anomaly, identifies the number of replayed signals, and determines the duration of the falsification all within a single unified framework.

It is not our intention to distinguish in depth between a replay attack and other abnormalities that seemingly may have the same effect e.g. a sensor malfunction or FDI, because these problems have been adequately addressed in relevant studies by using rule-based [33] or matrix-based distance-profile methods [34] for single sensor scenarios. Moreover, a direct comparison with classical techniques like watermarking or $x^2$ detectors is not feasible, as those methods assume prior knowledge of the compromised signal. Instead, our objective is to present an effective and robust framework capable of handling a sophisticated multisensory cyber anomaly scenario under time evolving reactor conditions, where the affected signals are not known a priori.

Despite its success, a limitation of the proposed framework is that the interpretation of Shapley values depends on the selection of appropriate background data. Existing studies typically use the global mean as the baseline, however, as explained, this assumption is not suitable for systems with strongly time dependent dynamics like nuclear reactors. To address this issue, we validated the approach during a dynamic reactor operation where the system evolution is known. Applying the method under steady state or different transient conditions would require using more representative baseline data. Future work will focus on developing a generalized solution. A promising direction is to deploy a Point Kinetics Equation (PKE) model of PUR-1 reactor in parallel with real time monitoring, enabling the model to continuously supply physics-based background data for the reactor's current operating condition.

Furthermore post hoc explanations using SHAP analysis variations have computational cost especially when instance-level explanations are extended to window-level. In the present work, the selected size of the process unit (10 seconds x 9 signals) is sufficient to provide real time performance without compromising interpretability. Future work will investigate how increasing the dimensions affects the scalability of the proposed framework.

Lastly, although the framework was benchmarked on real-world datasets from PUR-1, as an example of a critical environment where root cause interpretation is required in real-time, its design is general and applicable to other cyber-physical systems. The three modules of the methodology are domain-agnostic, which makes the framework suitable for applications in energy systems, robotics, smart-grid infrastructure and process control, provided that baseline data representing expected system behavior are available. In these domains, the framework can support sensor drift detection and localization, trajectory validation, actuator-level anomaly detection or cyber-induced perturbations.

## Acknowledgements

This research was performed using funding received from the DOE Office of Nuclear Energy's Nuclear Energy University Program under contract DE-NE0009268.